\documentclass[conference]{IEEEtran}
\usepackage{cite}

\ifCLASSINFOpdf
\else
\fi
\usepackage{amsmath}
\usepackage[ruled]{algorithm2e}
\usepackage{svg}
\usepackage{hhline}
\usepackage{balance}
\usepackage{adjustbox}
\usepackage{balance}
\usepackage{graphicx}
\usepackage{amssymb}
\usepackage{tikz}
\usetikzlibrary{patterns}

\newtheorem{assumption}{\textsc{Assumption}}

\begin{document}
%
\title{Online Unsupervised Multi-view Feature Selection}



%
\author{\IEEEauthorblockN{Weixiang Shao\IEEEauthorrefmark{1},
		Lifang He\IEEEauthorrefmark{2},
		Chun-Ta Lu\IEEEauthorrefmark{1}, 
		Xiaokai Wei\IEEEauthorrefmark{1} and
		Philip S. Yu\IEEEauthorrefmark{1}\IEEEauthorrefmark{3}}
	\IEEEauthorblockA{\IEEEauthorrefmark{1}University of Illinois at Chicago\\ 
		Email: \{wshao4, clu29, xwei2, psyu\}@uic.edu}
	\IEEEauthorblockA{\IEEEauthorrefmark{2}Shenzhen University, China\\
		Email: lifanghescut@gmail.com}
	\IEEEauthorblockA{\IEEEauthorrefmark{3}Institute for Data Science, Tsinghua University, China}
}


\maketitle

\begin{abstract}
 In the era of big data, it is becoming common to have data  with multiple modalities or coming from multiple sources, known as ``multi-view data''.
Since multi-view data are usually unlabeled and come from high-dimensional spaces (such as language vocabularies), 
unsupervised multi-view feature selection is crucial to many applications such as model interpretation and storage reduction.
However, it is nontrivial due to the following challenges. 
First, the data may not fit in memory, because there are too many instances or the feature dimensionality is too large. 
How to select useful features with limited memory space? 
Second, the data may come in as streams and concept drift may happen.
How to select features from streaming data and handles the concept drift? 
Third, different views may share some consistent and complementary information. 
How to leverage the consistent and complementary information from different views to improve the feature selection in the situation when the data are too big or come in as streams?
To the best of our knowledge, 
none of the previous works can solve all the challenges simultaneously. 

In this paper, we propose an Online unsupervised Multi-View Feature Selection, OMVFS, 
which deals with large-scale/streaming multi-view data in an online fashion.
OMVFS embeds unsupervised feature selection into a clustering algorithm via nonnegative matrix factorizatio with sparse learning.
It further incorporates the graph regularization to preserve the local structure information and help select discriminative features.
Instead of storing all the historical data, OMVFS processes the multi-view data chunk by chunk and aggregates all the necessary information into several small matrices. 
By using the buffering technique, the proposed OMVFS can reduce the computational and storage cost while taking advantage of the structure information. 
Furthermore, OMVFS can capture the concept drifts in the data streams.
Extensive experiments on four real-world datasets show the effectiveness and efficiency of the proposed OMVFS method. More importantly, OMVFS is about 100 times faster than the off-line methods.
\end{abstract}


%
\IEEEpeerreviewmaketitle

\section{Introduction}
In many real-world applications, data are often with multiple modalities or coming from multiple sources.
Such data are called multi-view data. 
For example, in web image retrieval, the visual information of images and their textual tags can be regarded as two views; 
in web document clustering, one web document may be translated into different languages, each language can be seen as one view.
Usually, multiple views provide complementary information for the semantically same data.
Multi-view learning was proposed to combine different views to obtain better performance than relying on just one single view \cite{sun2013survey}.
However, many of the views may come from high-dimensional spaces (such as language vocabularies).
Thus, accurate unsupervised feature selection on these high-dimensional multi-view  data is crucial to many applications such as model interpretation and storage reduction. 

Feature selection has been studied for decades \cite{john1994irrelevant}. 
It can be categorized into supervised feature selection and unsupervised feature selection in terms of the label availability.
Supervised feature selection \cite{zhao2007spectral} uses the class labels to
effectively select discriminative features to distinguish samples from different classes. 
As the data explodes, most of the data are unlabeled and expensive to obtain the labels.
Recently, several approaches on unsupervised feature selection has been proposed \cite{cai2010unsupervised,wang2015embedded}. 
Without the label information, most of the unsupervised feature selection methods combine generated pseudo labels with sparse learning \cite{li2012unsupervised,qian2013robust}.
In this paper, we will only focus on the unsupervised feature selection.

Most recently, as complementary information can be obtained from different views, unsupervised multi-view feature selection has drawn lots of attention \cite{wang2013multi,tang2013unsupervised,qian2014unsupervised}. 
For example, \cite{tang2013unsupervised} is the first to use spectral clustering and $ \ell_{2,1} $-norm regression
to multi-view data in social media.
\cite{wang2013multi} integrates all features and learns the weights for every feature with respect to each cluster individually via a new joint structured sparsity-inducing norms.
For image and text data, 
\cite{qian2014unsupervised} uses image local learning regularized orthogonal nonnegative matrix factorization to learn pseudo labels and simultaneously perform robust joint $ \ell_{2,1} $-norm minimization to select discriminative features.

However, several challenges prevent us from applying existing unsupervised feature selection methods to the real-world multi-view data:
\begin{enumerate}
    \item As information explodes, multi-view data may contain too many instances or the feature dimensionality may be too large, such that the data cannot fit in memory.
    How to select useful features with limited memory space is the first challenge.
	\item In many real-world applications, the data may come in as streams and concept drift \cite{wang2003mining} may happen.
	How to select features from streaming data and handles the concept drift is the second challenge.
	\item Multi-view data always exhibits the heterogeneity of the features.
	Different views  may share some consistent and complementary information. 
	The third challenge is how to combine features from different views while take advantages of the consistent and complementary information to improve the feature selection in the situation when the data are too big or come in as streams.
\end{enumerate}

To the best of our knowledge, all the existing feature selection methods only focus on one or two of the challenges.
For example, \cite{wang2013multi,tang2013unsupervised,qian2014unsupervised} only solve unsupervised feature selection problem for multi-view data.
\cite{song2013fast,maung2013pass,fsds} solve the problem of feature selection on large-scale/streaming data in a single view.
None of them can solve all the challenges simultaneously.

In this paper, we propose a novel method, Online unsupervised Multi-View Feature Selection (OMVFS), to solve the problem of multi-view feature selection on large-scale/streaming data.
OMVFS embeds the unsupervised feature selection into the Nonnegative Matrix Factorization (NMF) based clustering objective function. 
It further adopts the graph regularization to preserve the local structure information and help select discriminative features.
By learning a consensus clustering indicator matrix, the proposed OMVFS integrates all the views in different feature spaces.
Instead of storing all the historical data, OMVFS processes the multi-view data chunk by chunk and aggregates necessary information from all the previous data into several small matrices.
These aggregated matrices will be used in the learning of feature selection matrices and be updated as new data come in.
The contributions of this paper can be summarized as following:
\begin{enumerate}
	\item The proposed OMVFS method is the first attempt to solve the problem of online unsupervised multi-view  feature selection on large-scale/streaming data.
	\item OMVFS can process the data chunks in an online fashion and aggregate information about all the previous data into small matrices. 
	The aggregated information can be used to help feature selection in the future. 
	Thus, the proposed method can greatly reduce the memory requirement and scale up to large data without appreciable sacrifice of performance.
	\item 
	By using the buffering technique, OMVFS can reduce the computational and storage cost while taking advantage of the structure information.
	Furthermore, it can capture the concept drifts in the data streams.
    \item Through extensive experiments on real-world datasets, we demonstrate that the effectiveness of proposed OMVFS is comparative and even better than the best off-line method. More importantly, OMVFS is about 100 times faster than the off-line methods.
\end{enumerate}
The rest of this paper is organized as follows. 
In the next section, problem formulation and some backgrounds are given. 
The details of the proposed OMVFS method are presented in Sections \ref{sec:method}, and \ref{sec:omvfs}. 
Extensive experimental results and analysis are shown in Section \ref{sec:experiments}. 
Related work is discussed in Section \ref{sec:related} and followed by the conclusion in Section \ref{sec:conclusion}.

\section{Preliminaries}
In this section, we will briefly describe the problem of unsupervised multi-view feature selection. 
Then some background knowledge about unsupervised feature selection will be introduced.

\subsection{Problem Description}
Before we describe the formulation of the problem, we summarize some notations used in this paper in Table \ref{tab:notations}.

Throughout this paper, matrices are written as boldface capital letters (\textit{e.g.}, $ \mathbf{M}\in \mathbb{R}^{n\times m} $) 
and vectors are denoted as boldface lowercase letters (\textit{e.g.}, $ \mathbf{m}_i $). 
$ \|\cdot\|_F $ is the matrix Frobenius norm and $ Tr(\cdot) $ is the trace of a square matrix.
The $ \ell_{2,1} $-norm is defined as $\|\mathbf{M}\|_{2,1} = \sum_{i=1}^{n}\mathbf{m}_i = \sum_{i=1}^{n}\sqrt{\sum_{j=1}^{m}\mathbf{M}_{i,j}^2}$.

Assume we are given a dataset with $ N $ instances in $ n_v $ views $ \{\mathbf{X}^{(v)}, v = 1,2,...,n_v\}$, 
where $ \mathbf{X}^{(v)} \in \mathbb{R}_+^{N\times D_v}$ represents the nonnegative data in the $ v $-th view 
and $ D_v $ is the feature dimension in the $ v $-th view.
Our goal is to leverage complementary information from multi-view data to select $ p_v $ features from the $ v $-th view, while dealing with high-dimensional and large-scale problems in an online fashion. 

\begin{table}[t]
	\centering
	\caption{Summary of the Notations}
	\label{tab:notations}
	\begin{adjustbox}{max width=\columnwidth}
		\begin{tabular}{l|l}
			Notation&Description\\
			\hline
			$ N $ & Total number of instances.\\
			$ D_v $ & Dimension of features for view $ v $.\\
			$ n_v $ & Total number of views.\\
		    $ m $ & Size of data received at each time.\\
			$ s $ & Size of the buffer.\\
			$ \mathbf{X}^{(v)}, \mathbf{W}^{(v)},  \mathbf{V}^{(v)}$ & Data matrix, similarity matrix, and feature selection matrix for view $ v $.\\
			$ \mathbf{U}$ & Cluster indicator matrix.\\
            $ \mathbf{M}_{t} $ & Matrix for time $t$.\\
			$ \mathbf{M}_{[t]} $ & Aggregated matrix from time $1$ to time $ t $.\\
			$ \mathbf{M}_{[s,t]} $ & Aggregated matrix from time $t-s+1$ to time $ t $.\\

		\end{tabular}
	\end{adjustbox}
\end{table}
\subsection{Unsupervised Feature Selection using NMF}
\label{sec:nmf}
Nonnegative Matrix Factorization (NMF) has been widely used in unsupervised learning such as clustering and dimension reduction.
In this paper, we embed the feature selection into a NMF based clustering algorithm.
Let $ \mathbf{X} \in \mathbb{R}_+^{N\times D}$ denote a nonnegative data matrix, 
where each row represents an instance and each column represents one normalized attribute (each column $ \|\mathbf{X}_{.,i}\|_2 =1$). 
Assume we would like to cluster the data into $ K $ clusters, 
NMF will factorize the data matrix $ \mathbf{X} $ into two nonnegative matrices. 
We denote the two nonnegative matrices factors as $ \mathbf{U}\in \mathbb{R}_+^{N\times K} $ and $ \mathbf{V}\in \mathbb{R}_+^{D\times K} $.
The objective function for NMF can be formulated as below:
\begin{equation}
\begin{split}
\min_{\mathbf{U},\mathbf{V}}~ \mathcal{L} = \|\mathbf{X} - \mathbf{U}\mathbf{V}^T\|_F^2\\
s.t.~\mathbf{U}\ge 0, \mathbf{V}\ge 0.
\end{split}
\label{nmf}
\end{equation}
In practice, for clustering problem, an orthogonality constraint is usually added to $ \mathbf{U} $ \cite{ding2006orthogonal,choi2008algorithms,yoo2010nonnegative}.
Thus, Eq.~(\ref{nmf}) can be rewritten as:
\begin{equation}
\begin{split}
\min_{\mathbf{U},\mathbf{V}}~ \|\mathbf{X} - \mathbf{U}\mathbf{V}^T\|_F^2\\
s.t.~\mathbf{U}^T\mathbf{U} = \mathbf{I}, \mathbf{U}\ge 0 , \mathbf{V}\ge 0.
\end{split}
\label{eq:orth-nmf}
\end{equation}
The orthogonality constraint on $ \mathbf{U} $ can be seen as a relaxed form from the original clustering indicator constraint, where $ \mathbf{U}\in\{0,1\}^{N\times K}, \mathbf{U}^T\mathbf{1} = \mathbf{1} $.

Another advantage of this orthogonality constraint on $ \mathbf{U} $ is 
to allow us to perform feature selection using $ \mathbf{V} $.
It has been proved that the $ \ell_2 $-norm of the rows in $ \mathbf{V} $ represents the importance of the features in $ \mathbf{X} $ regarding to the reconstruction error \cite{EUFS}.
Thus, we can add a selection matrix $ diag(\mathbf{p}) $ to $ \mathbf{X} $ and $ \mathbf{V} $ to only select the important features that minimize the reconstruction error:,
\begin{equation}
\begin{split}
\min_{\mathbf{U},\mathbf{V}}~ &\|\mathbf{X}diag(\mathbf{p}) - \mathbf{U}(diag(\mathbf{p})\mathbf{V})^T\|_F^2 \\
s.t.~&\mathbf{U}^T\mathbf{U} = \mathbf{I}, \mathbf{U}\ge 0 , \mathbf{V}\ge 0\\
 &\mathbf{p}\in \{0,1\}^{D}, \mathbf{p}^T\mathbf{1} = r,
\end{split}
\label{eq:mixed_integer}
\end{equation}
where $ r $ is the number of selected feature, $ \mathbf{p} $ is the indicator vector, where $ p_i=1 $ indicates that the $ i $-th feature is selected. 

Now, Eq.~(\ref{eq:mixed_integer}) becomes a mixed integer programming, which is very difficult to solve.
\cite{EUFS} proposed to relax Eq.~(\ref{eq:mixed_integer}) to the following problem and proved the equivalent between Eq.~(\ref{eq:mixed_integer}) and Eq.~(\ref{eq:mixed_integer2}): 
\begin{equation}
\begin{split}
\min_{\mathbf{U},\mathbf{V}}~ &\|\mathbf{X} - \mathbf{U}\mathbf{V}^T\|_F^2 +  \beta \|\mathbf{V}\|_{2,1}  \\
s.t.~&\mathbf{U}^T\mathbf{U} = \mathbf{I}, \mathbf{U}\ge 0 , \mathbf{V}\ge 0,\\
\end{split}
\label{eq:mixed_integer2}
\end{equation}
where $ \beta $ is the parameter to control the sparsity of $ \mathbf{V} $.
By adding the $ \ell_{2,1} $-norm on $ \mathbf{V} $, we force some of the rows in $ \mathbf{V} $ close to $ 0 $.
Thus, we can achieve feature selection by sorting the features according to the row norms of $ \mathbf{V} $ in descending order, and selecting the top ranked ones.
To facilitate discussions, we call $ \mathbf{U} $ the \textit{\textbf{cluster indicator matrix}}, and $ \mathbf{V} $ the \textit{\textbf{feature selection matrix}}.


\section{Method}
\label{sec:method}
The proposed online unsupervised multi-view feature selection is based on nonnegative matrix factorization 
and it processes the multi-view data chunk by chunk and aggregates all the historical information into small matrices with low computational and storage complexity. 
We will first describe how to derive the objective function. 

\subsection{Objective of OMVFS} 
Given data in $ n_v $ views $ \{\mathbf{X}^{(v)}\in \mathbb{R}_+^{N\times D_v}, v = 1,2,...,n_v\} $, 
we aim to find a feature selection matrix for each view and a consensus cluster indicator matrix, 
which integrates information of all the views.
Following the constrained NMF framework in Section \ref{sec:nmf}, we can form the objective function
\begin{equation}
\begin{adjustbox}{max width=0.91\columnwidth}
$\displaystyle
\begin{split}
\min_{\mathbf{U},\{\mathbf{V}^{(v)}\}} &\sum_{v=1}^{n_v} \left( \| \mathbf{X}^{(v)} - \mathbf{U}{\mathbf{V}^{(v)}}^T \|_F^2 + \beta_v \|\mathbf{V}^{(v)}\|_{2,1} \right)\\
&\textit{s.t.}~~ \mathbf{U}^T\mathbf{U} = \mathbf{I}, \mathbf{U} \ge 0,  \mathbf{V}^{(v)} \ge 0, v = 1,2,...,n_v,\\
\end{split}
$
\end{adjustbox}
\end{equation}
where $ \mathbf{U} $ is the consensus cluster indicator matrix, $ \mathbf{V}^{(v)} $ is the feature selection matrix for the $ v $-th view, 
and $ \beta_v $ is the parameter that controls the sparsity of $ \mathbf{V}^{(v)} $.

To take advantage of local manifold information from the structure of the original data $ \{\mathbf{X}^{(v)}\} $, 
\textit{i.e.}, similar data instances should have similar labels, 
we add the spectral clustering objective term for every view
\begin{equation}
\min Tr({\mathbf{U}^{(v)}}^T\mathbf{L}^{(v)}\mathbf{U}^{(v)})
\end{equation}
where $ \mathbf{L}^{(v)} = \mathbf{D_w}^{(v)} - \mathbf{W}^{(v)} $ is the Laplacian matrix for the $ v $-th view,
$ \mathbf{W}^{(v)} \in \mathbb{R}^{N\times N}$ is the similarity matrix based on $ \mathbf{X}^{(v)} $ 
and $ \mathbf{D_w}^{(v)} $ is a diagonal matrix with its diagonal elements as the row sums of $ \mathbf{W}^{(v)} $.
Now, we can obtain the objective function for OMVFS as:
\begin{equation}
\begin{adjustbox}{max width=0.91\columnwidth}
$\displaystyle
\begin{split}
\min_{\mathbf{U},\{\mathbf{V}^{(v)}\}} &\sum_{v=1}^{n_v} \left( \| \mathbf{X}^{(v)} - \mathbf{U}{\mathbf{V}^{(v)}}^T \|_F^2 + \alpha_v tr(\mathbf{U}^T \mathbf{L}^{(v)}\mathbf{U}) + \beta_v \|\mathbf{V}^{(v)}\|_{2,1} \right)\\
&\textit{s.t.}~~ \mathbf{U}^T\mathbf{U} = \mathbf{I}, \mathbf{U} \ge 0,  \mathbf{V}^{(v)} \ge 0, v = 1,2,...,n_v\\
\end{split}
$
\end{adjustbox}
\label{eq:offline}
\end{equation}
where $ \alpha_v $ is the importance of the spectral clustering term for the $ v $-th view.

It is worth noting that in order to solve the problem in Eq.~(\ref{eq:offline}), 
we need to have the entire data $ \{\mathbf{X}^{(v)}\} $ in memory.
However, in real-world applications, the data may be too large to fit into the memory or may only come on stream.
Hence, it is crucial to solve the above optimization problem in an incremental way.
Let $ \mathbf{X}^{(v)}_t \in \mathbb{R}_+^{m\times D_v} $ denote the data received at time $ t $ in the $ v $-th view, 
and $ \mathbf{X}^{(v)}_{[t]} \in \mathbb{R}_+^{mt\times D_v}$ denote all the data received up to time $ t $,
where $ m $ is the number of instances (size of the data chunk) received at each time.
The objective function at time $ t $ is:
\begin{equation}
\begin{adjustbox}{max width=0.91\columnwidth}
$\displaystyle
\begin{split}
\min_{\mathbf{U}_{[t]},\{\mathbf{V}^{(v)}\}} &\sum_{v=1}^{n_v} \left( \| \mathbf{X}^{(v)}_{[t]} - \mathbf{U}_{[t]}{\mathbf{V}^{(v)}}^T \|_F^2 + \alpha_v Tr(\mathbf{U}_{[t]}^T \mathbf{L}_{[t]}^{(v)}\mathbf{U}_{[t]}) \right) \\
& + \sum_{v=1}^{n_v} \beta_v \|\mathbf{V}^{(v)}\|_{2,1} \\
\textit{s.t.}~~& \mathbf{U_{[t]}}^T\mathbf{U}_{[t]} = \mathbf{I}, \mathbf{U}_{[t]} \ge 0,  \mathbf{V}^{(v)} \ge 0, v = 1,2,...,n_v\\
\end{split}
$
\end{adjustbox}
\label{eq:online}
\end{equation}
where $ \mathbf{U}_{[t]} \in \mathbb{R}_+^{mt\times K}$ is the consensus cluster indicator matrix for all the instances received up to time $ t $ and 
$ \mathbf{L}^{(v)}_{[t]} \in \mathbb{R}^{mt\times mt}$ is the Laplacian matrix constructed from $ \mathbf{X}^{(v)}_{[t]} $.
\subsection{Optimization}
\label{sec:optimization}
In the previous section, we derived the objective function of OMVFS. 
To solve OMVFS, we first rewrite the optimization problem as follows
\begin{equation}
\begin{adjustbox}{max width=0.91\columnwidth}
$\displaystyle
\begin{split}
\min_{\mathbf{U}_{[t]},\{\mathbf{V}^{(v)}\}} &\sum_{v=1}^{n_v} \left( \| \mathbf{X}^{(v)}_{[t]} - \mathbf{U}_{[t]}{\mathbf{V}^{(v)}}^T \|_F^2 + \alpha_v Tr(\mathbf{U}_{[t]}^T \mathbf{L}_{[t]}^{(v)}\mathbf{U}_{[t]})\right) \\
+& \sum_{v=1}^{n_v} \beta_v \|\mathbf{V}^{(v)}\|_{2,1} + \gamma \|\mathbf{U}_{[t]}^T\mathbf{U}_{[t]}-\mathbf{I} \|_F^2\\
&\textit{s.t.}~~ \mathbf{U}_{[t]} \ge 0,  \mathbf{V}^{(v)} \ge 0, v = 1,2,...,n_v\\
\end{split}
$
\end{adjustbox}
\label{eq:online2}
\end{equation}
where $ \gamma > 0 $ is a parameter to control the orthogonality condition.
In practice, $ \gamma $ should be large enough to ensure the orthogonality is satisfied ($\gamma $ is set to $ 10^7 $ throughout the experiments).
From Eq.~(\ref{eq:online2}), we can see that at each time $ t $, we need to optimize $ \mathbf{U}_{[t]} $ and $ \{\mathbf{V}^{(v)}\}_{v=1}^{n_v} $.
However, the objective function is not jointly convex, so we have to update  $ \mathbf{U}_{[t]} $ and $ \{\mathbf{V}^{(v)}\}_{v=1}^{n_v} $
in an alternating way. 
Thus, there are two subproblems in OMVFS:

\subsubsection{Optimize $ \mathbf{U}_{[t]} $ with $ \{\mathbf{V}^{(v)}\}_{v=1}^{n_v} $ Fixed}
To optimize $ \mathbf{U}_{[t]} $ with $ \{\mathbf{V}^{(v)}\}_{v=1}^{n_v} $ fixed at time $ t $,
we only need to minimize the following objective:
\begin{equation}
\begin{adjustbox}{max width=0.98\columnwidth}
$\displaystyle
\begin{split}
\mathcal{J}_t(\mathbf{U}_{[t]}) =&\sum_{v=1}^{n_v} \left( \| \mathbf{X}^{(v)}_{[t]} - \mathbf{U}_{[t]}{\mathbf{V}^{(v)}}^T \|_F^2 + \alpha_v tr(\mathbf{U}_{[t]}^T \mathbf{L}_{[t]}^{(v)}\mathbf{U}_{[t]})\right)\\
&+ \gamma \|\mathbf{U}_{[t]}^T\mathbf{U}_{[t]}-\mathbf{I} \|_F^2\\
&\textit{s.t.}~~\mathbf{U}_{[t]} \ge 0\\
\end{split}
$
\end{adjustbox}
\label{eq:u_sofa}
\end{equation}

Taking the first-order derivative, the gradient of $ \mathcal{J}_t $ with respect to $ \mathbf{U}_{[t]} $ is
\begin{equation}
\begin{adjustbox}{max width=0.8\columnwidth}
$\displaystyle
\begin{split}
\frac{\partial \mathcal{J}_t}{\partial \mathbf{U}_{[t]}} =& 2\mathbf{U}_{[t]}\sum_{v=1}^{n_v}{\mathbf{V}^{(v)}}^T\mathbf{V}^{(v)} 
-2\sum_{v=1}^{n_v}\mathbf{X}_{[t]}^{(v)}\mathbf{V}_{[t]}^{(v)}\\ 
&+ 2\mathbf{M}_{[t]}\mathbf{U}_{[t]} + 2\gamma\mathbf{T}_{[t]} - 2\gamma\mathbf{U}_{[t]}
\end{split}
$
\end{adjustbox}
\label{eq:grad_u}
\end{equation}
where, $ \mathbf{M}_{[t]} =\sum_{v=1}^{n_v}\alpha_v\mathbf{L}^{(v)}_{[t]} $ and $ \mathbf{T}_{[t]} = \mathbf{U}_{[t]}\mathbf{U}_{[t]}^T\mathbf{U}_{[t]} $.

Using the Karush-Kuhn-Tucker (KKT) complementary condition for the nonnegativity constraint on $ \mathbf{U}_{[t]} $, 
we can get the update rule for $ \mathbf{U}_{[t]} $ \cite{boyd2004convex,choi2008algorithms}:
\begin{equation}
\begin{adjustbox}{max width=0.98\columnwidth}
$\displaystyle
\begin{split}
(\mathbf{U}_{[t]})_{i,k} \leftarrow (\mathbf{U}_{[t]})_{i,k}\sqrt{
	\frac{(\sum_{v=1}^{n_v}\mathbf{X}_{[t]}^{(v)}\mathbf{V}^{(v)} + \lambda\mathbf{U}_{[t]} + \mathbf{M}_{[t]}^-\mathbf{U}_{[t]})_{i,k}}
	{(\mathbf{U}_{[t]}\sum_{v=1}^{n_v}{\mathbf{V}^{(v)}}^T\mathbf{V}^{(v)} + \gamma\mathbf{T}_{[t]} + \mathbf{M}_{[t]}^+\mathbf{U}_{[t]})_{i,k}}
}
\end{split}
$
\end{adjustbox}
\label{eq:update_u_all}
\end{equation}
where $ (\mathbf{M}_{[t]}^+)_{i,j} = \frac{1}{2}\left(\|(\mathbf{M}_{[t]})_{i,j}\| + (\mathbf{M}_{[t]})_{i,j}\right) $,
$ (\mathbf{M}_{[t]}^-)_{i,j} = $ \\$\frac{1}{2}\left(\|(\mathbf{M}_{[t]})_{i,j}\| - (\mathbf{M}_{[t]})_{i,j}\right) $ and 
$ \mathbf{M}_{[t]} = \mathbf{M}_{[t]}^+ - \mathbf{M}_{[t]}^- $.

\subsubsection{Optimize $ \{\mathbf{V}^{(v)}\}_{v=1}^{n_v} $ with $ \mathbf{U}_{[t]} $ Fixed}
From Eq.~(\ref{eq:online2}), we can observe that the optimization of $ \mathbf{V}^{(v)} $ is independent with different $ v $ when $ \mathbf{U}_{[t]} $ is fixed.
We only need to minimize the following objective function for the $ v $-th view:
\begin{equation}
\begin{adjustbox}{max width=0.91\columnwidth}
$\displaystyle
\begin{split}
\mathcal{J}_t(\mathbf{V}^{(v)}) &= \| \mathbf{X}_{[t]}^{(v)} - \mathbf{U}_{[t]}{\mathbf{V}^{(v)}}^T \|_F^2 + \beta_v \|\mathbf{V}^{(v)}\|_{2,1}\\
&\textit{s.t.}~~\mathbf{V}^{(v)} \ge 0\\
\end{split}
$
\label{eq:v_object}
\end{adjustbox}
\end{equation}
We observe that the above objective $ \mathcal{J}_t $ can be decomposed as:
\begin{equation}
\begin{split}
\mathcal{J}_t(\mathbf{V}^{(v)}) &= \sum_{i=1}^{t}\| \mathbf{X}_{i}^{(v)} - \mathbf{U}_{i}{\mathbf{V}^{(v)}}^T \|_F^2 + \beta_v \|\mathbf{V}^{(v)}\|_{2,1}\\
&\textit{s.t.}~~\mathbf{V}^{(v)} \ge 0
\end{split}
\end{equation}
where $ \mathbf{X}_{i}^{(v)} $ is the data chunk received at time $ i $ and $ \mathbf{U}_{i} $ is the cluster indicator matrix for data received at time $ i $.

Taking the first-order derivative, the gradient of $ \mathcal{J}_t $ with respect to $ \mathbf{V}^{(v)} $ is
\begin{equation}
\frac{\partial \mathcal{J}_t}{\partial \mathbf{V}^{(v)}} = 2\mathbf{V}^{(v)}\sum_{i=1}^{t}\mathbf{U}_i^T\mathbf{U}_i - 2\sum_{i=1}^{t}{\mathbf{X}_i^{(v)}}^T\mathbf{U}_i 
+ \beta_v\mathbf{D}^{(v)}\mathbf{V}^{(v)}
\label{eq:grad_v}
\end{equation}
where $ \mathbf{D}^{(v)} $ is a diagonal matrix with the $ j $-th diagonal element given by $\mathbf{D}^{(v)}_{j,j} = \frac{1}{\|\mathbf{v}^{(v)}_j\|_2}$
and $ \mathbf{v}^{(v)}_j $ is the $ j $-th row of $ \mathbf{V}^{(v)} $ and $ \|\cdot\|_2 $ is the $ \ell_2 $ norm.

For the sake of convenience, we introduce two terms $ \mathbf{A}_t $ and $ \mathbf{B}^{(v)}_t $:
{
	\begin{align}
	\mathbf{A}_t &= \sum_{i=1}^{t}\mathbf{U}_i^T\mathbf{U}_i = \mathbf{A}_{t-1}+\mathbf{U}_t^T\mathbf{U}_t\\
	\mathbf{B}^{(v)}_t &= \sum_{i=1}^{t}{\mathbf{X}_i^{(v)}}^T\mathbf{U}_i = \mathbf{B}^{(v)}_{t-1} + {\mathbf{X}_t^{(v)}}^T\mathbf{U}_t
	\end{align}
}
Both $ \mathbf{A}_t $ and $ \mathbf{B}^{(v)}_t $ can be computed incrementally with low storage. 
Thus, Eq.~\ref{eq:grad_v} can be rewritten as
\begin{equation}
\begin{split}
\frac{\partial \mathcal{J}_t}{\partial \mathbf{V}^{(v)}} =& 2\mathbf{V}^{(v)}(\mathbf{A}_{t-1}+\mathbf{U}_t^T\mathbf{U}_t) \\
&- 2(\mathbf{B}^{(v)}_{t-1} + {\mathbf{X}_t^{(v)}}^T\mathbf{U}_t) + \beta_v\mathbf{D}^{(v)}\mathbf{V}^{(v)}
\end{split}
\end{equation}
Using the KKT complementary condition for the nonnegativity constraint on $ \mathbf{V}^{(v)} $, 
we get the update rule for $ \mathbf{V}^{(v)} $:
\begin{equation}
\begin{adjustbox}{max width=0.90\columnwidth}
$\displaystyle
\mathbf{V}^{(v)}_{j,k} \leftarrow \mathbf{V}^{(v)}_{j,k} \sqrt{
	\frac{\left(\mathbf{B}^{(v)}_{t-1} + {\mathbf{X}_t^{(v)}}^T\mathbf{U}_t\right)_{j,k}}
	{\left(\mathbf{V}^{(v)}(\mathbf{A}_{t-1}+\mathbf{U}_t^T\mathbf{U}_t) + \frac{1}{2}\beta_v\mathbf{D}^{(v)}\mathbf{V}^{(v)}\right)_{j,k}} }
$
\end{adjustbox}
\label{eq:update_v}
\end{equation}
\subsection{Further Optimization Using Buffering}
From the update rules (\ref{eq:update_u_all}), 
we observe that the update process for $ \mathbf{U}_{[t]} $ still needs all the data and all the cluster indicator matrices up to time $ t $ to reside in the memory.
Thus, the memory usage will continue growing as more data come in.
Also, at each time $ t $, we need to calculate the new similarity matrix $ \mathbf{W}_{[t]}^{(v)} $ in order to get the Laplacian matrix $ \mathbf{L}_{[t]}^{(v)} $.
Even if we only calculate the similarity between the $ m $ new instances and all the $ mt $ instances, 
the time complexity will still increase as the data pile up.
What's worse, the size of $ \mathbf{L}_{[t]}^{(v)} $ ($ mt\times mt $) will grow quadratically.
To overcome this deficiency, we adopt the buffering technique by keeping a limited number of samples for each update step. 
Another benefit of adopting buffering is that, by only considering the structure information in the buffer, we can capture the concept drift in streaming data.
The intuition behind buffering is based on the assumption regarding to streaming data:

\begin{assumption}
	\label{theo:convergence}
	The incoming data are more related with the recent data than the far more old data, 
	and the underline concept distribution of incoming data are more similar to those of recent data.
\end{assumption}

Thus, we consider the graph/structure information only within a time window. 
Assume that the buffer size is $ s $,  \textit{i.e.}, we only consider the most recent $ s $ data chunks received, 
we define $ \mathbf{X}^{(v)}_{[s,t]} = [{\mathbf{X}^{(v)}_{t-s+1}};{\mathbf{X}^{(v)}_{t-s+2}}; \dots; {\mathbf{X}^{(v)}_{t}}] \in \mathbb{R}_+^{sm\times D_v} $ as the most recent $ s $ data chunks received up to time $ t $ and $ \mathbf{U}_{[s,t]} = [\mathbf{U}_{t-s+1};\mathbf{U}_{t-s+2};\dots;\mathbf{U}_{t}] \in \mathbb{R}_+^{sm\times K} $ as the cluster indicator matrix for them.
$ \mathbf{W}_{[s,t]}^{(v)} \in \mathbb{R}^{sm \times sm}$ and $ \mathbf{L}^{(v)}_{[s,t]} \in \mathbb{R}^{sm \times sm}$ are the similarity matrix constructed from the data in the buffer and the Laplacian matrix from $ \mathbf{W}^{(v)}_{[s,t]} $.
Note that, for the similarity matrix $ \mathbf{W}_{[s,t]}^{(v)} $, 
we do not need to recompute the similarities among all the $ sm $ instances ($ sm\times sm $ pairwise similarities).
We only need to calculate the similarities between the $ m $ new instances with all the $ sm $ instances ($ m\times sm $ pairwise similarities).
Thus, we can redefine the objective function (\ref{eq:u_sofa}) by substituting $ \mathbf{X}^{(v)}_{[t]} $, $ \mathbf{U}_{[t]} $ and $ \mathbf{L}^{(v)}_{[t]} $ with the above new definitions.
The new update rule for $ \mathbf{U}_{[s,t]} $ is
\begin{equation}
\begin{adjustbox}{max width=0.99\columnwidth}
$\displaystyle
\begin{split}
(\mathbf{U}_{[s,t]})_{i,k} \leftarrow (\mathbf{U}_{[s,t]})_{i,k}\sqrt{
	\frac{(\sum_{v=1}^{n_v}\mathbf{X}_{[s,t]}^{(v)}\mathbf{V}^{(v)} + \lambda\mathbf{U}_{[s,t]} + \mathbf{M}_{[s,t]}^-\mathbf{U}_{[s,t]})_{i,k}}
	{(\mathbf{U}_{[s,t]}\sum_{v=1}^{n_v}{\mathbf{V}^{(v)}}^T\mathbf{V}^{(v)} + \gamma\mathbf{T}_{[s,t]} + \mathbf{M}_{[s,t]}^+\mathbf{U}_{[s,t]})_{i,k}}
}
\end{split}
$
\end{adjustbox}
\label{eq:update_u_buff}
\end{equation}

For the feature selection matrices $ \{\mathbf{V}^{(v)}\}_{v=1}^{n_v} $, we fortunately observe that 
$ \mathbf{A}_t $ and $\{ \mathbf{B}^{(v)}_t\}_{v=1}^{n_v}  $ accumulate the information of all the previous $ t $ chunks. 
Therefore, we can still use update rule (\ref{eq:update_v}) without any modification,
which means that employing the buffering technique has no effect on updating the feature selection matrices.

\section{Online Unsupervised Multi-View Feature Selection}
\label{sec:omvfs}
\begin{figure}
	\centering
	\includegraphics[width=\columnwidth]{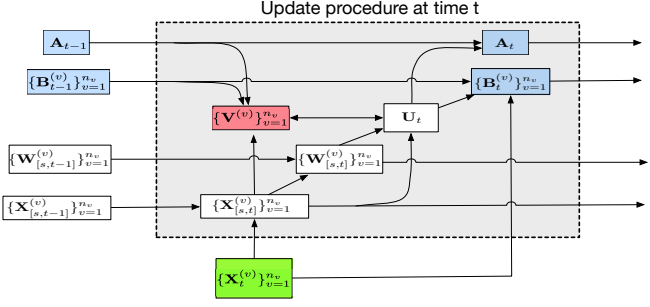}
	\caption{The workflow of OMVFS at time $ t $. The blue boxes represent the aggregated information about all the previously received data, the green box represents the incoming data at time $ t $, the red box represents the feature weight matrices learned from the aggregated information and the new data.}
	\label{fig:flow}
\end{figure}

In this section we will present the whole OMVFS method. 
Fig.~\ref{fig:flow} shows the workflow of the OMVFS method at time $t$.
From Fig.~\ref{fig:flow} we can see that, at time $t$, 
OMVFS takes $\mathbf{A}_{t-1} $, $\{\mathbf{B}_{t-1}^{(v)}\}_{v=1}^{n_v}$, $\mathbf{W}_{[s,t-1]}$, and $\{\mathbf{X}^{v}_{[s,t-1]}\}^{n_v}_{v=1}$ from time $t-1$ and combines with the incoming data $\{\mathbf{X}^{v}_{t}\}^{n_v}_{v=1}$ to update all the matrices.
It is worth mentioning that OMVFS does not need to store all the previous data.
Instead, it only stores some small buffered data $\{\mathbf{X}^{v}_{[s,t-1]}\}^{n_v}_{v=1}$ and $\mathbf{W}_{[s,t-1]}$.
All the necessary information from the previous data is aggregated and stored in small matrices $\mathbf{A}_{t-1} $ and $\{\mathbf{B}_{t-1}^{(v)}\}_{v=1}^{n_v}$.
The computation of these aggregated matrices can be done efficiently and incrementally.
The complete algorithm procedure is shown in Algorithm \ref{algorithm_OMVFS}.
There are several things need to be clarified. 

\begin{figure}
	\centering
	\includegraphics[width=\columnwidth]{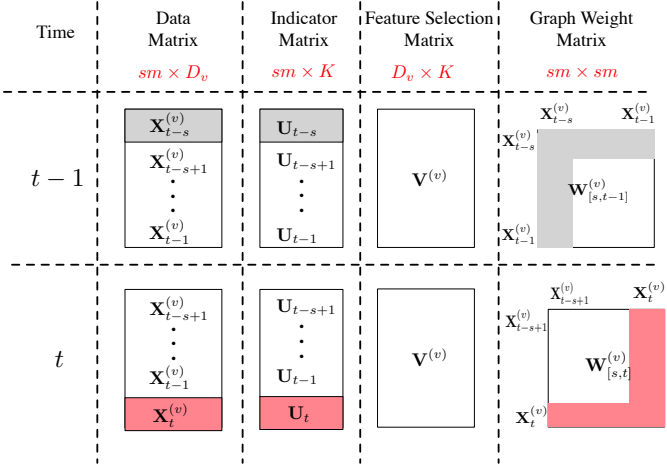}
	\caption{The detailed update process for major matrices in OMVFS at time $ t $. 
	When the buffer is full, data for the oldest chunk (gray parts) will be discarded, and data for the incoming chunck (red parts) will be added. }
	\label{fig:update}
\end{figure}

First, at the beginning of the algorithm ($ t = 0 $), OMVFS will initialize the buffered data $ \mathbf{X}^{(v)}_{[s,t]} $, 
the cluster indicator matrix for buffered data $ \mathbf{U}_{[s,t]} $ and similarity matrix for buffered data $ \mathbf{W}_{[s,t]}^{(v)} $ with empty matrices.
The feature selection matrices $ \{\mathbf{V}^{(v)}\} $ will also be initialized randomly at $ t = 0 $.

Second, as the data come in, if the buffer is full, 
OMVFS will remove the oldest data (\textit{e.g.}, $ \mathbf{X}^{(v)}_{t-s} $),
and concatenate the new data to the buffer.
The procedure for updating the matrices in the buffer is shown in Fig.~\ref{fig:update}.
The grey shadow areas are the parts for the oldest data, and the red shadow areas are the parts for the incoming new data.

\begin{algorithm}[t]                      
	\caption{OMVFS algorithm.}          
	\label{algorithm_OMVFS}                           
	\small
	\LinesNumbered \SetAlgoVlined
	\KwIn{Data matrices $\{\mathbf{X}^{(v)}\}$.
	The number of clusters $ K $, the batch size $ m $, the buffer size $ s $.
	Parameters $ \{\alpha_v\} $ and $\{\beta_v\}$.}
	\KwOut{Feature selection matrices $ \{\mathbf{V}^{v}\} $.}
	\BlankLine
	Initialize $ \mathbf{V}^{(v)} $ randomly for each view $ v $. \\
	Initialize $ \mathbf{X}^{(v)}_{[s,0]} $, $ \mathbf{U}_{[s,0]} $, $ \mathbf{W}^{(v)}_{[s,0]} $ as empty matrices. \\
	$ \mathbf{A}^{(v)}_0  = \textbf{0},~ \mathbf{B}^{(v)}_0 = \textbf{0}$ for each view $ v $. \\
	\For{$ t=1:\lceil N/m \rceil $}{
	    Draw $ \mathbf{X}^{(v)}_t $ for all the views.\\
	    Initialize $ \mathbf{U}_t $ randomly.\\
	
		\For{$ v = 1:n_v $}{
		Construct $ \mathbf{X}^{(v)}_{[s,t]} $ from $ \mathbf{X}^{(v)}_{[s,t-1]} $.\\
		Construct $ \mathbf{U}_{[s,t]} $ from $ \mathbf{U}_{[s,t-1]} $.\\
		Construct  $ \mathbf{W}^{(v)}_{[s,t]} $ and $ \mathbf{L}^{(v)}_{[s,t]} $ from $ \mathbf{W}^{(v)}_{[s,t-1]} $.\\
    	}
		\Repeat{Convergence}{
		    Update $ \mathbf{U}_{[s,t]} $ according to Eq.~(\ref{eq:update_u_buff}).\\
    		\For{$ v = 1:n_v $}{
	    		Update $ \mathbf{V}^{(v)} $ according to Eq.~(\ref{eq:update_v}).\\
		    }
	    }
	    $ \mathbf{A}_t = \mathbf{A}_{t-1}+\mathbf{U}_t^T\mathbf{U}_t$ \\
	    $ \mathbf{B}^{(v)}_t = \mathbf{B}^{(v)}_{t-1} + {\mathbf{X}_t^{(v)}}^T\mathbf{U}_t $\\
	}
	\For{$ v = 1:n_v $}{
	    Sort the features for $ \mathbf{X}^{(v)} $ according to the $ \ell_2 $-norm of the rows in $ \mathbf{V}^{(v)} $ in descending order.\\
	}
\end{algorithm}

\subsection{Convergence Analysis}
Although the objective function for the proposed OMVFS at time $ t $ is not jointly convex with $ \mathbf{U}_{[s,t]} $ and $ \{\mathbf{V}^{(v)}\}$, 
the alternating optimization strategy used in OMVFS is guaranteed to converge to the local minimum \cite{lee2001algorithms}. The proof is similar to that of Theorem 1 in \cite{cai2011graph} and we omit it. It is important to note that our method essentially applies stochastic gradient descent \cite{bottou2010large} to the new coming data. Since we perform a stochastic approximation for minimizing an objective function that is written as a sum of differentiable functions. This method is expected to decrease the objective function in every iteration on the new data \cite{bottou1998online}. Therefore, the same convergence proof can be adapted to the problem setting and algorithm design that considered here. In the experiments, we also find that the proposed OMVFS method converges within 200 iterations for all the datasets used.


\subsection{Complexity Analysis}
There are two subproblems for OMVFS algorithm: optimizing $ \mathbf{U}_{[s,t]} $, and optimizing $ \{\mathbf{V}^{(v)}\} $. 
The computation cost for updating $ \mathbf{U}_{[s,t]} $ depends on the calculation in Eq.~(\ref{eq:update_u_buff}).
After analyzing the equation, we can find that the computational cost to update $ \mathbf{U}_{[s,t]} $ is $ O(n_vsmDK) + O(n_vs^2m^2K) $,
where $ s $ is the buffer size, $ m $ is the size of data chunk, 
and $ D $ is the average feature dimension for all the views.
Since $ D$ is usually very large, we can assume that $ D \gg sm $. 
Thus, the computational cost for updating $ \mathbf{U}_{[s,t]} $ is $ O(n_vsmDK) $.

The compuational cost for updating $ \mathbf{V}^{(v)} $ depends on Eq.~(\ref{eq:update_v}).
The computation for $ \mathbf{D}^{(v)} $ is $ O(D_vK) $.
Since $ \mathbf{D}^{(v)} $ is diagonal, it only takes $ O(D_vK) $ to calculate $ \mathbf{D}^{(v)}\mathbf{V}^{(v)} $. 
It is easy to verify that it take $ O(mD_vK) $ to update $ \mathbf{V}^{(v)} $, and $ O(n_vmDK) $ to update $ \{\mathbf{V}^{(v)}\} $.
Thus, the time complexity for OMVFS to process one data chunk is $ O(tn_vsmDK) $, 
which leads to the overall time complexity to $ O(tn_vsNDK) $.
Here, $ t $ is the average number of iterations to converge.
Considering $ s $ is usually very small (less than 5 in the experiments), the time complexity is approximately $ O(tn_vNDK) $.
We can observe that the time complexity is linear to all the components (number of views, number of instances, feature dimensionality and the number of clusters).

Most of the offline methods require at least $ O(n_vND) $ space, however the proposed OMVFS only requires $ O(n_vsmD) + O(n_vs^2m^2)\approx O(n_vsmD) $ memory space, which makes OMVFS suitable for large-scale/streaming data.

\section{Experiments and Results}
\label{sec:experiments}
\subsection{Dataset}
In this paper, two small real-world datasets and two large real-world datasets are used to evaluate the proposed OMVFS method.
The summary of these four datasets is shown in Table \ref{tab:data}, 
and the details of the datasets are as follows:
\begin{itemize}
	\item \textbf{CNN} and \textbf{FOX}\footnote{https://sites.google.com/site/qianmingjie/home/datasets/}:
	These two datasets were crawled from CNN and FOX web news \cite{qian2014unsupervised}. 
	The category information contained in the RSS feeds for each news article can be viewed as reliable class label.
	Each instance can be represented in two views, the text view and image view.
	Titles, abstracts, and text body contents are extracted as the text view data, and the image associated with the article is stored as the image view data.
	All text content is stemmed by portStemmer, and l2-normalized TF-IDF is used as text features, which results in $ 35,719 $ features for CNN and $ 27,072 $ features for FOX.
	For image features, seven groups of color features and five textural features are used \cite{qian2014unsupervised}, which results in $ 996 $ features for both datasets.
	\item \textbf{YouTube Multiview Video Games (YouTube)}\footnote{https://archive.ics.uci.edu/ml/datasets/YouTube+Multiview\newline+Video+Games+Dataset}:
	This dataset consists of feature values and class labels for about 120,000 videos (instances) \cite{madaniEtAl2013MLJ}. 
	Each instance is described by up to 13 feature types, 
	from 3 high level feature families: textual, visual, and auditory features. 
	There are 31 class labels, 1 through 31. 
	The first 30 labels correspond to popular video games.
	Class 31 is not specific, and means none of the 30.
	In this paper, we select one text view and one audio view for the experiments.
	For the text view, we use the $ 94,012 $ unigrams extracted from the video tags.
	We use the Mel-Frequency Cepstral Coefficient (MFCC) features with the dimension of $ 2,000 $ for the audio view.
	We removed some instances with label 31 and kept $ 95,343 $ instances to create a more balanced dataset. 
	\item \textbf{Reuters Multilingual Text Data (Reuters)}\footnote{http://archive.ics.uci.edu/ml/machine-learning-databases/00259/}:
	The text collection contains feature characteristics of documents originally written in five different languages (English, French, German, Spanish and Italian), and their translations, over a common set of 6 topic categories \cite{Amini09learningfrom}.
	In our experiments we use the $ 111,740 $ documents in two languages (English and French).
	The feature dimensions are $ 21,531 $ for English and $ 24,893 $ for French.
\end{itemize}
\begin{table}[t]
	\centering
	\caption{Summary of the datasets}
	\label{tab:data}
	\begin{tabular}{|c|c|c|c|}
		\hline
		Dataset & \# Instance & \# Feature & \# Class\\
		\hline
		CNN & $ 2,107 $ & $ 996 + 35,719 $ &7\\
		FOX & $ 1,523 $ & $ 996 + 27,072$ & 4\\
		YouTube & $ 95,343 $ & $ 2,000 + 94,012 $ & 31 \\
		Reuters & $ 111,740 $ & $ 21,531 + 24,893 $ & 6\\
		\hline
	\end{tabular}
\end{table}
\subsection{Comparison Methods}
We compare the proposed OMVC method with several state-of-art methods. 
The differences between these comparison methods are summarized in Table \ref{tab:methods},
and the details of comparison methods are as follows:
\begin{itemize}
	\item \textbf{OMVFS}: OMVFS is the online unsupervised multi-view feature selection method proposed in this paper \footnote{code available at: https://github.com/software-shao}.
	
	\item \textbf{LapScore}: Laplacian Score \cite{he2005laplacian} is a single view unsupervised feature selection method, which evaluates the importance of a feature via its power of locality preservation. 
	\item \textbf{EUFS}: Embedded Unsupervised Feature Selection \cite{EUFS} is one of the most recent single view unsupervised feature selection method. 
	It directly embeds unsupervised feature selection algorithm into a clustering algorithm via sparse learning.
	\item \textbf{FSDS}: Unsupervised Feature Selection on Data Stream \cite{fsds} is one of the state-of-the-art unsupervised feature selection algorithms, which handles large volume single view data stream efficiently. 
	It adopts the idea of matrix sketching to efficiently maintain a low-rank approximation of the observed data and applies regularized regression to select the important features.
	\item \textbf{MVUFS} Multi-View Unsupervised Feature Selection \cite{mvufs} is the most advanced off-line unsupervised feature selection for multi-view data. 
	It uses local learning regularized orthogonal nonnegative matrix factorization to learn pseudo labels 
	and simultaneously performs joint $ \ell_{2,1} $-norm minimization to select discriminative features.
\end{itemize}
\begin{table}[t]
	\centering
	\caption{Summary of the comparison methods}
	\label{tab:methods}
	\begin{adjustbox}{max width=\columnwidth}
		\begin{tabular}{|c|c|c|c|}
			\hline
			Methods & Multi-view & Online & Direct Embedding \\
			\hline
			LapScore& $ \times $ & $ \times $ & $ - $\\
			EUFS & $ \times $ & $ \times $ &  \checkmark  \\
			MVUFS & \checkmark & $ \times $ & $ \times $ \\ 
			FSDS & $ \times $ & \checkmark & $ \times $ \\ 
			OMVFS & \checkmark & \checkmark & \checkmark \\ 
			\hline
		\end{tabular}
	\end{adjustbox}
\end{table}

\subsection{Experiment Settings}
In our experiments, two widely used evaluation metrics, Accuracy (ACC) and Normalized Mutual Information (NMI), 
are used to measure the clustering performance \cite{DBLP:conf/sigir/XuLG03}. 
We apply different methods to the four data sets, 
then the multi-view spherical K-means algorithm \cite{mvc} is applied to get the clustering solution. 
Since LapScore, EUFS and FSDS only work for single view data,
in the experiments, we apply these three methods on each of the view to select features.
It is worth mentioning that MVUFS is an off-line multi-view feature selection method, 
which takse all data into consideration 
and can often achieve better performance than online methods. 
Furthermore, LapScore, EUFS, MVUFS are all off-line methods, which cannot handle large datasets.
Thus, only the two online methods, FSDS and OMVFS, are applied to the two large datasets.

Most of the comparison methods use graph/similarity matrices.
In the experiments, we used the same kernal/similarity matrices as stated in the original papers.
For the proposed OMVFS, we used the Gaussian kernel.
If not stated, for the online methods (FSDS and OMVFS) the size of data chunk is set to 200 
for the two small datasets and 1000 for the two large datasets.
The buffer size is set to 2  and $ \gamma $ is set to $ 10^7 $ for the proposed OMVFS method.
For the sake of convenience, the parameters $ \alpha_v $ and  $ \beta_v $ are all set equally for different views.
For FSDS, we do grid search in $ \{1,2,...,10\} $ for the index of parameter $ \alpha $.
For all the other parameters in the comparison methods, we do grid search in $ \{10^{-2},10^{-1},...,10^{2}\} $.
We also vary the number of selected features as $ \{100,200,...,600\} $ for all the views.
The performance of the best parameter setting is reported for all the methods.

\subsection{Results on Small Datasets}
\begin{figure}
	\centering
	\subfloat[ACC for FOX]{\includegraphics[width= .49\columnwidth]{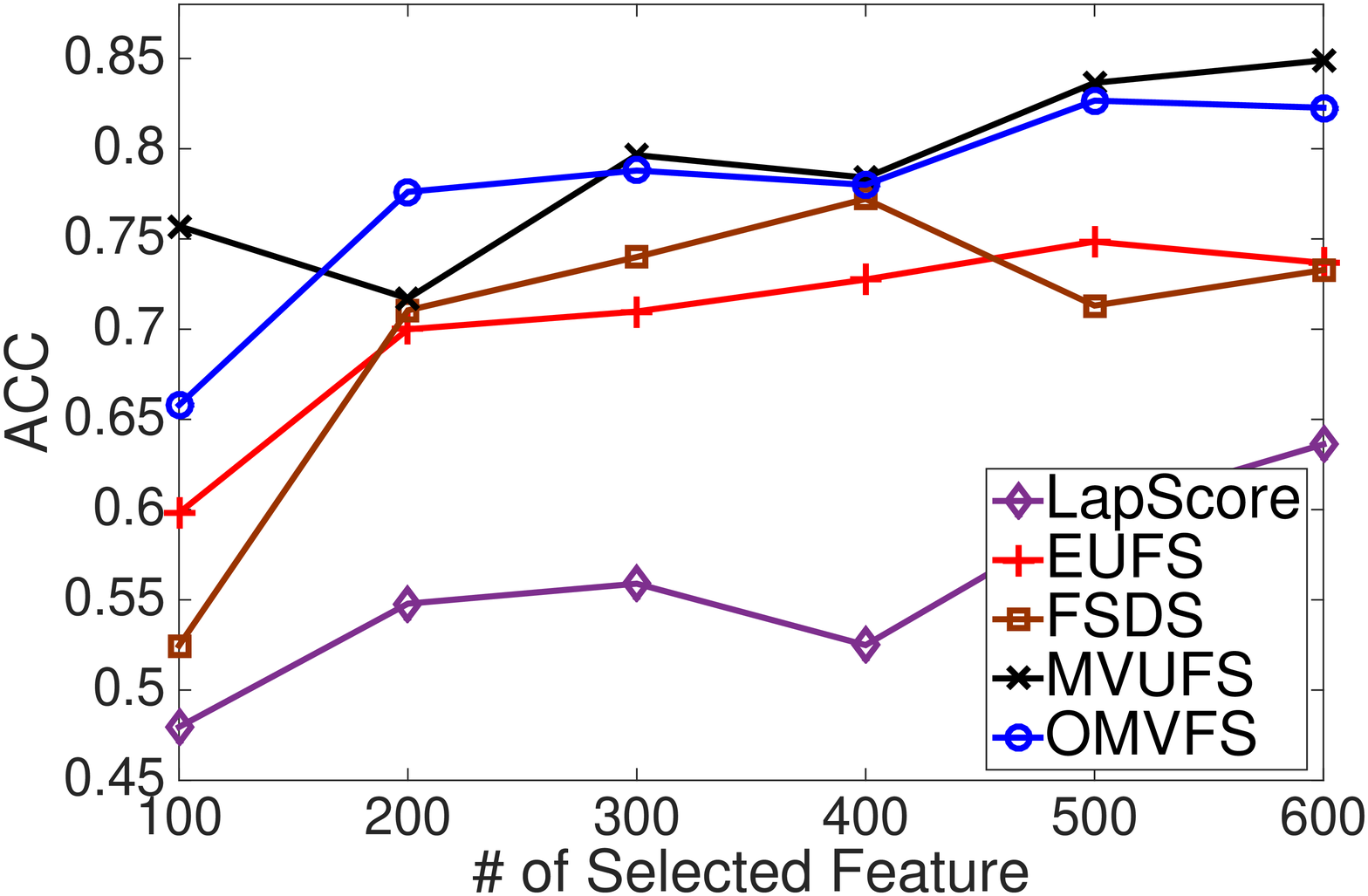}\label{fig:fox_a}}
	~ 
	\subfloat[NMI for FOX]{\includegraphics[width= .49\columnwidth]{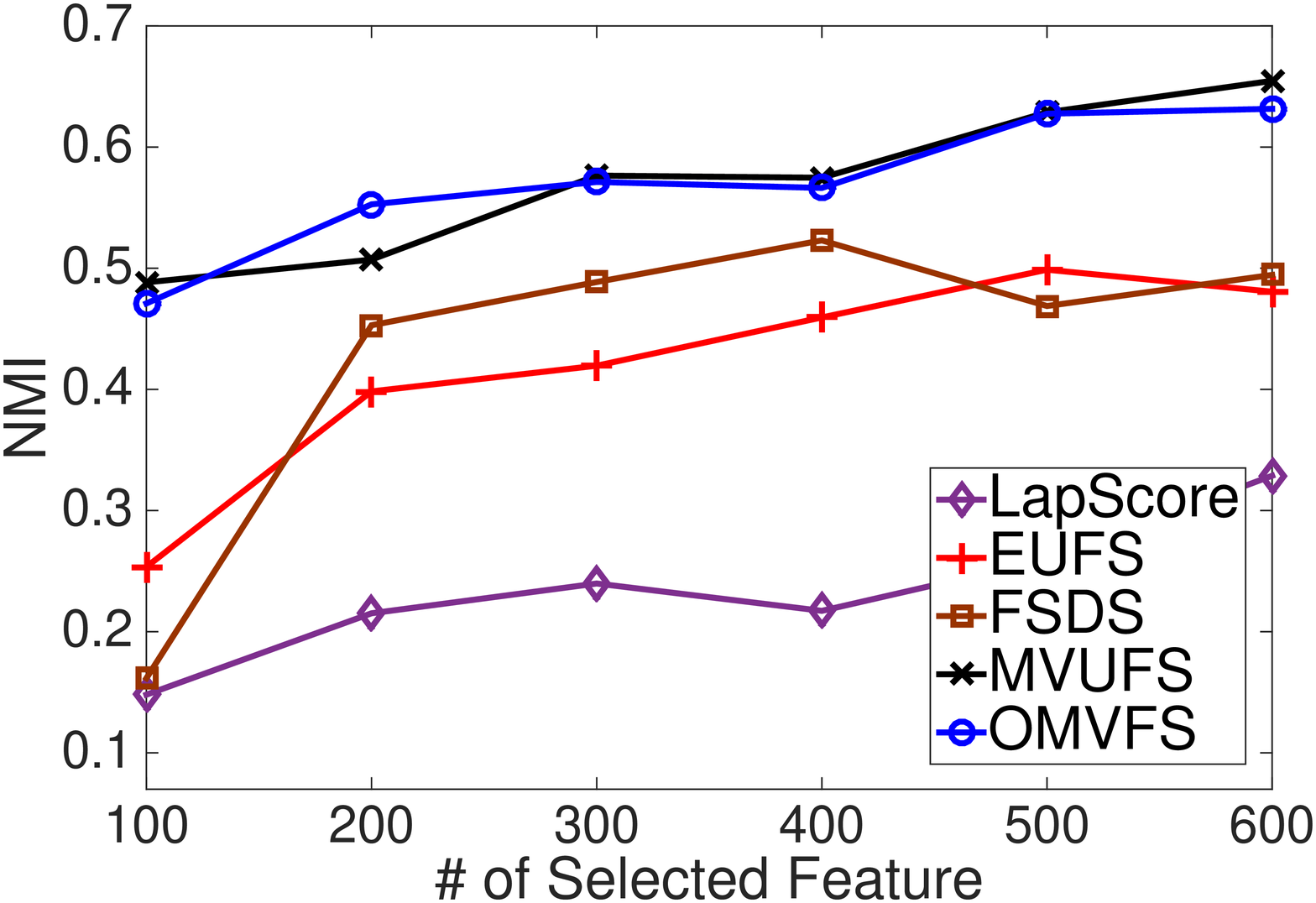}\label{fig:fox_b}}
	\\
	\subfloat[ACC for CNN]{\includegraphics[width= .49\columnwidth]{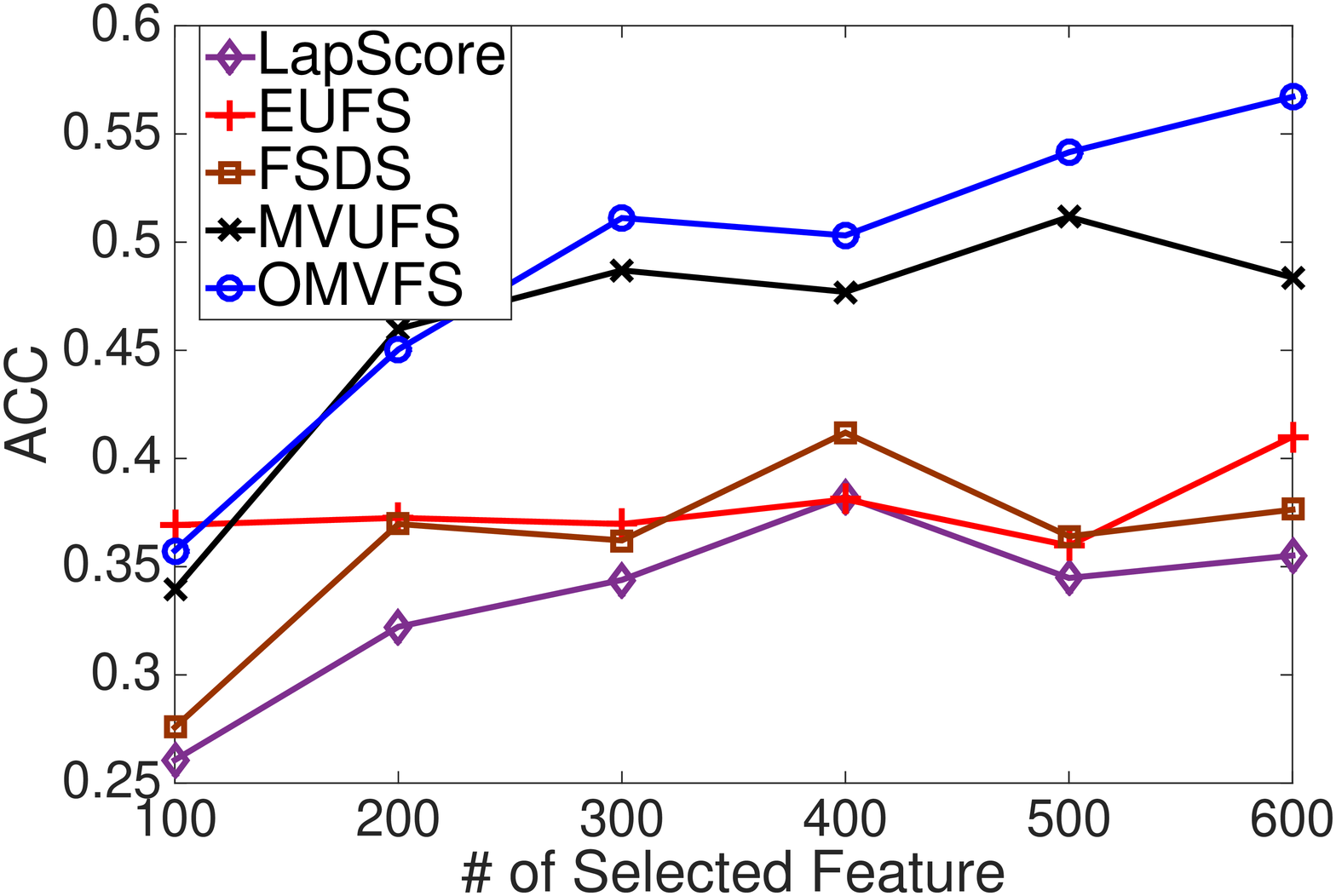}}
	~
	\subfloat[NMI for CNN]{\includegraphics[width= .49\columnwidth]{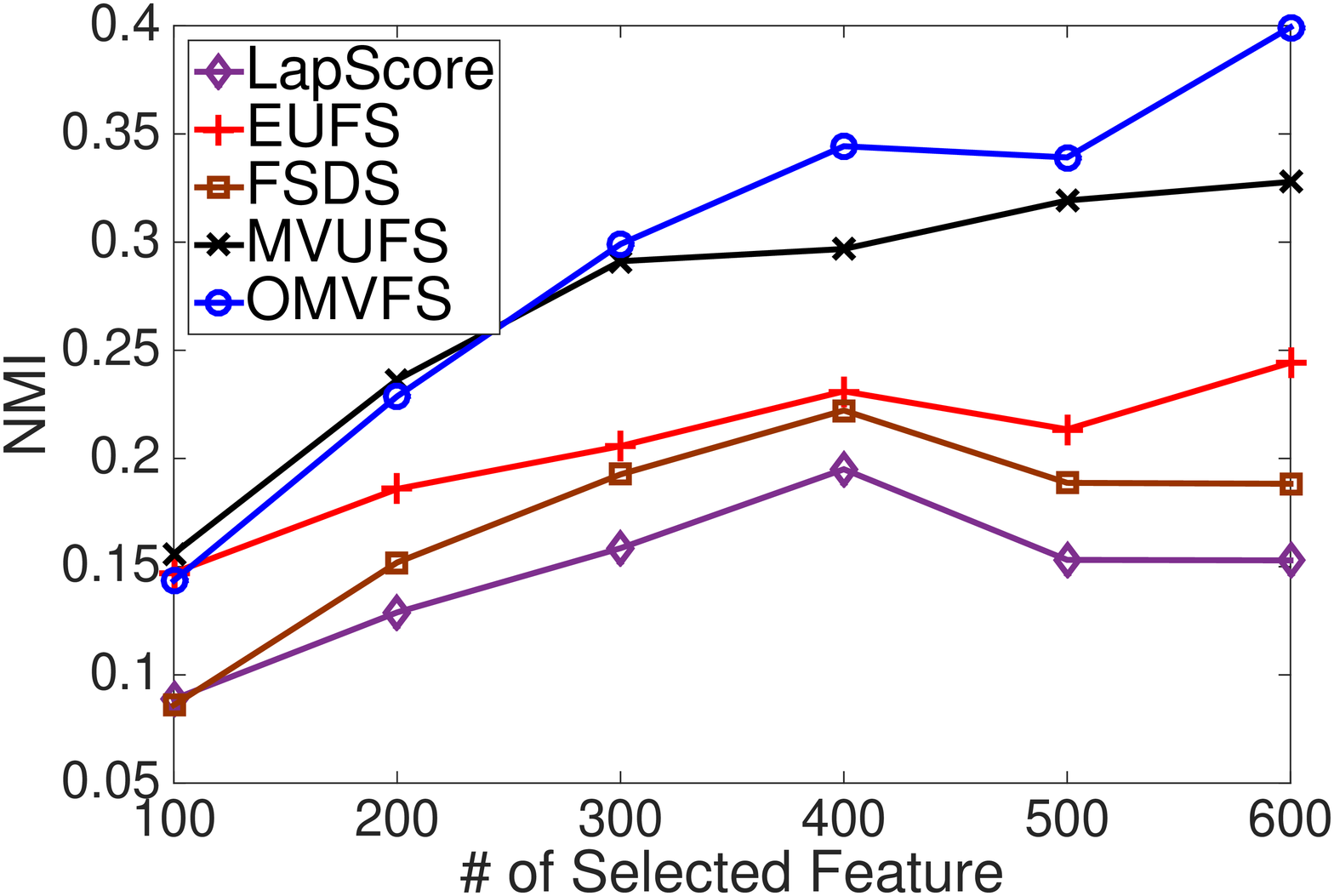}}
	\caption{Performance with different \# features on small datasets.}
	\label{fig:small}
\end{figure}

In order to compare the proposed OMVFS method with other comparison methods, we first apply all the methods to two small datasets, FOX and CNN.
Fig.~\ref{fig:small} shows the performance of all the methods on FOX and CNN data with different numbers of selected features.

From the Fig.~\ref{fig:small}, 
we can observe that as the number of selected feature increases, 
the performance will increase for most of the cases.
For both datasets, the two multi-view feature selection methods, MVUFS (off-line) and OMVFS (online),
outperform the single view feature selection methods.
This observation suggests that combining different views would benefit the feature selection for multi-view data.

For the FOX data, we can observe from Fig.~\ref{fig:fox_a} and Fig.~\ref{fig:fox_b} that
the performance of the proposed OMVFS is very close and even better than that of the best off-line method.
Among the three single view methods, FSDS and EUFS can get better performance than LapScore.
Although FSDS may suffer from low performance when including only 100 features,
its performance will increase dramatically when including more than 200 features.
This may indicate that the online FSDS selects less discriminative features as the top features,
however, more discriminative features will be included if we increase the size of the feature set.

For the CNN data, we can see that the proposed OMVFS has similar performance as MVUFS when the feature size is less than 300.
However, OMVFS outperforms MVUFS if more than 300 features are included. 
Comparing the statics of CNN and FOX data, we can find that CNN has higher feature dimensions and more classes than FOX.
The proposed OMVFS may select more discriminative features in data with higher dimensions than MVUFS.

From the results on FOX and CNN, 
we can conclude that the proposed OMVFS method outperforms all the single view feature selection methods (online and off-line).
It can also get close or even better performance than the most advanced off-line multi-view feature selection method.

\subsection{Result on Large Datasets}
\begin{figure}
	\centering
	\subfloat[ACC for Reuters]{\includegraphics[width= .49\columnwidth]{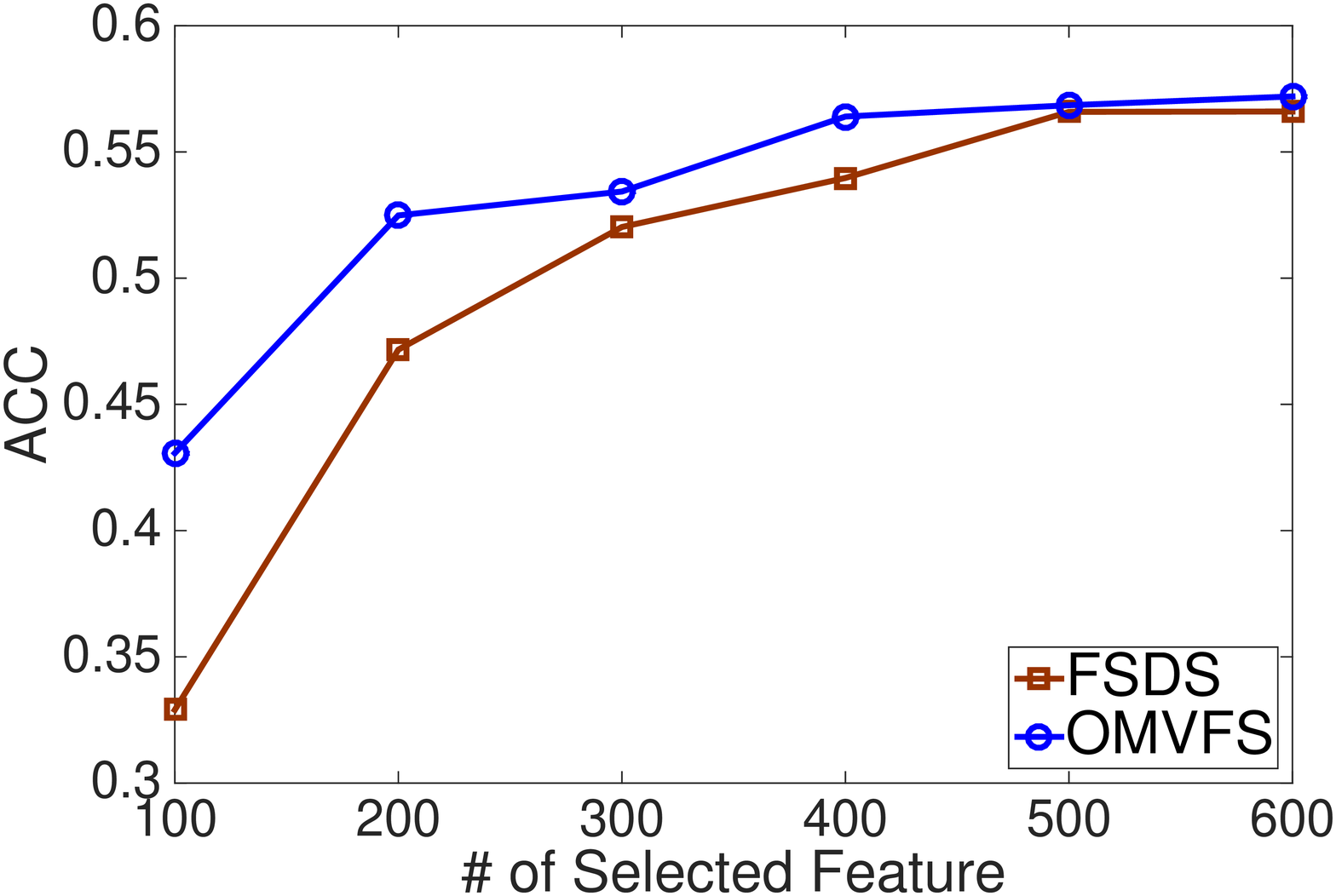}}
	~ 
	\subfloat[NMI for Reuters]{\includegraphics[width= .49\columnwidth]{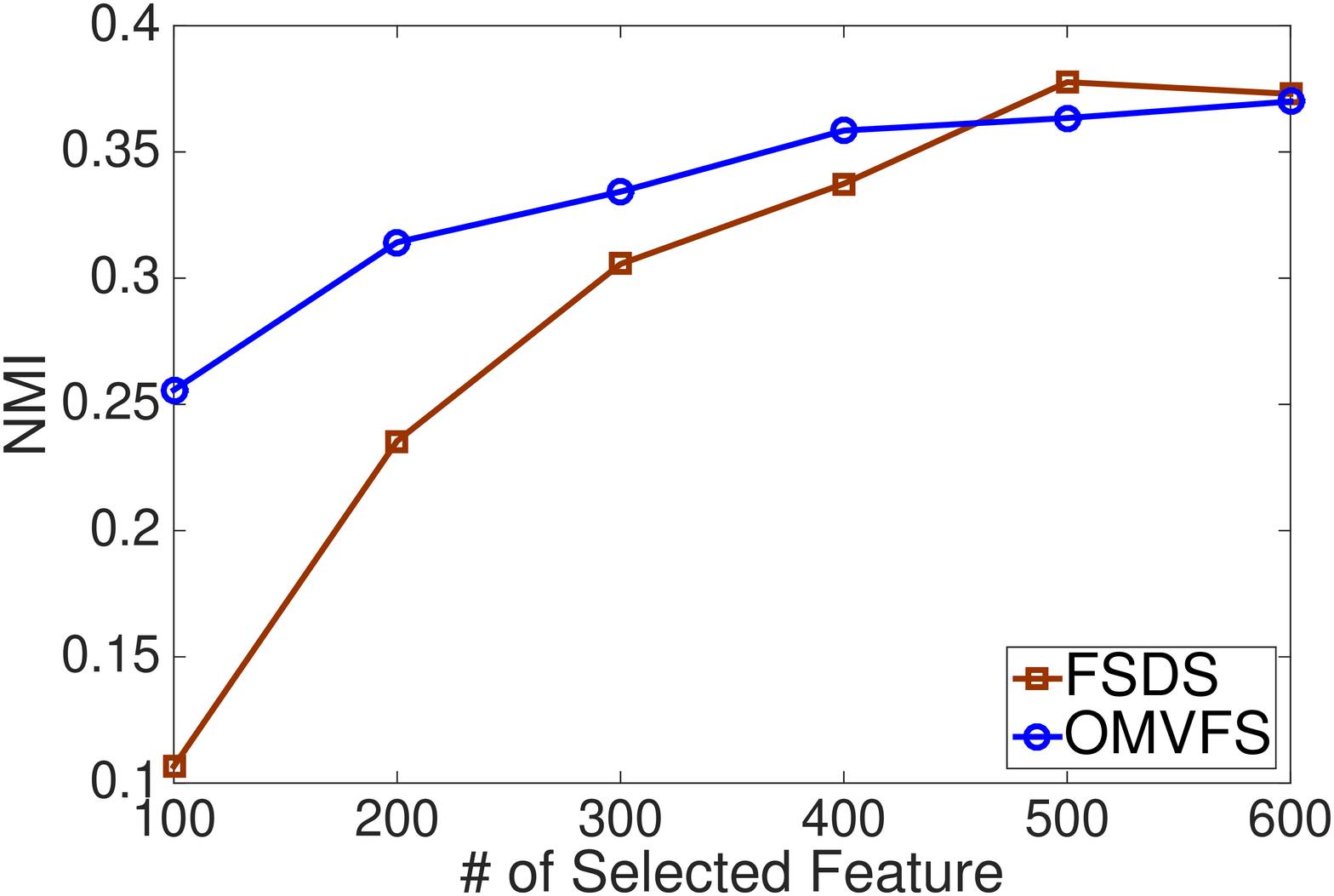}}
	\\
	\subfloat[ACC for YouTube]{\includegraphics[width= .49\columnwidth]{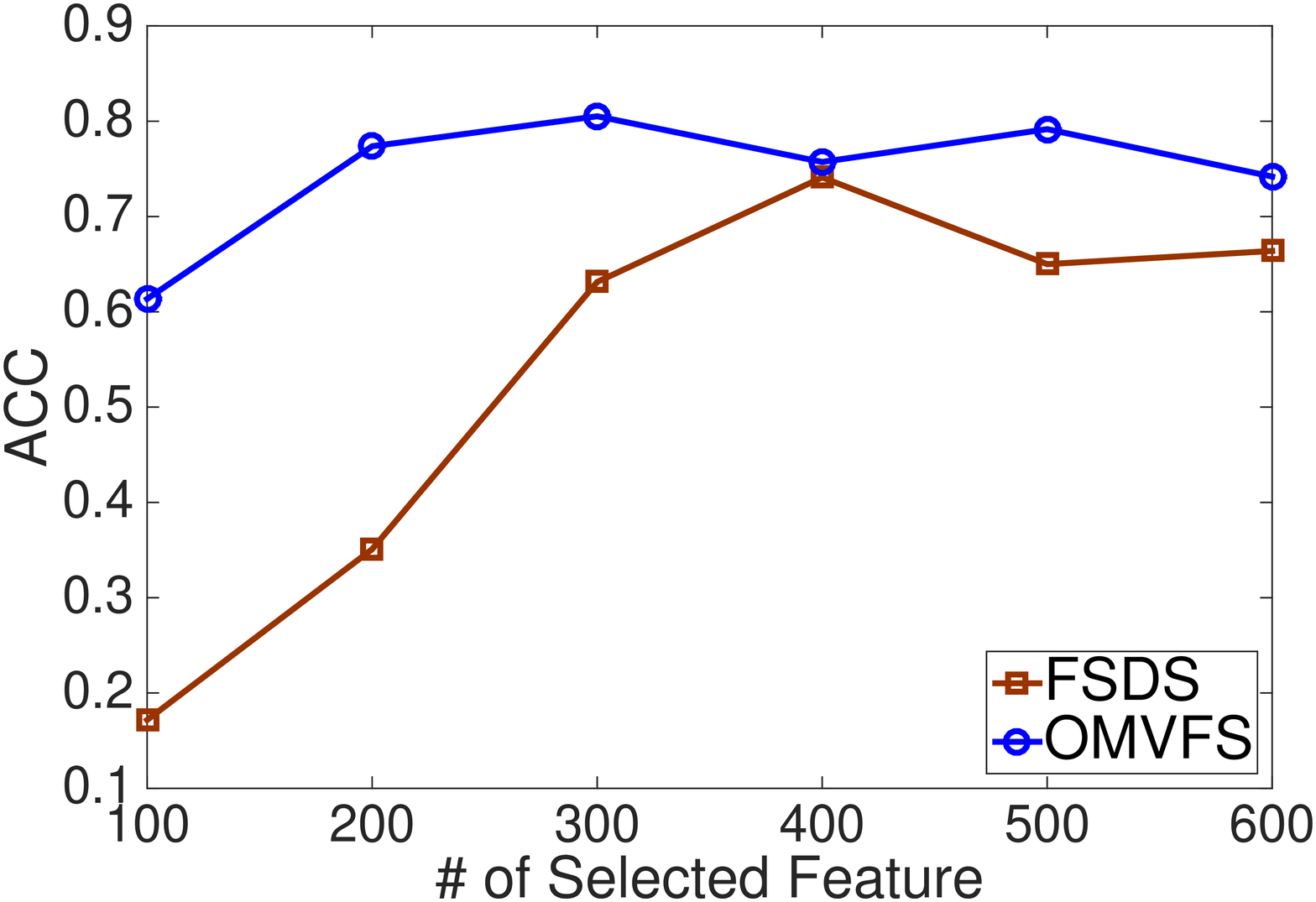}}
	~
	\subfloat[NMI for YouTube]{\includegraphics[width= .49\columnwidth]{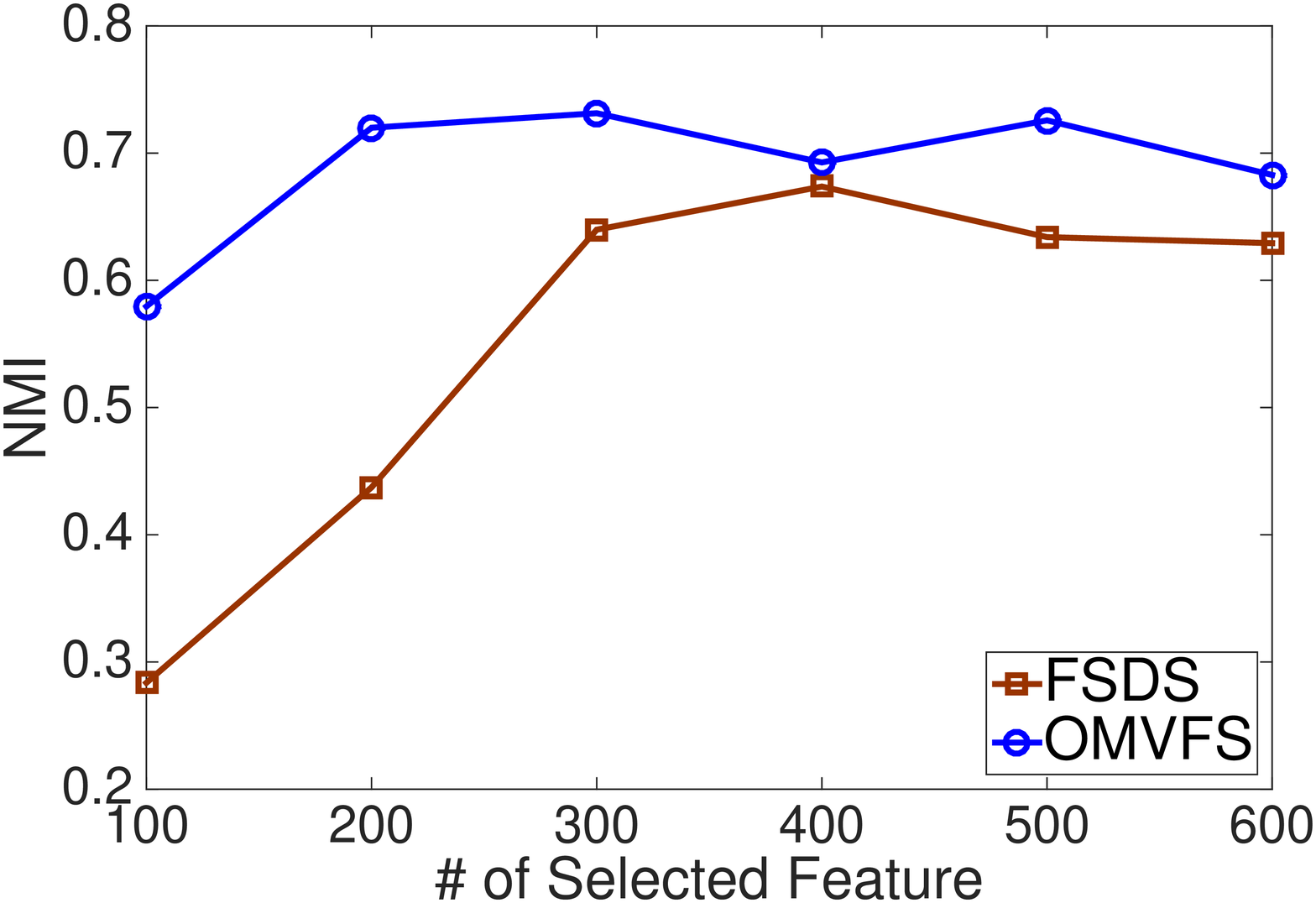}}
	\caption{Performance with different \# features on large datasets.}
	\label{fig:large}
\end{figure}

Since the proposed OMVFS is designed for large-scale multi-view datasets, 
we test OMVFS on the two largest public available multi-view datasets.
We also compare OMVFS with the most recent online single view feature selection method, FSDS.
The results on both datasets are reported in Fig.~\ref{fig:large}.
However, due to the extremely high runtime and memory consumption, 
LapScore, EUFS and MVUFS cannot be applied to the two large datasets.

From Fig.~\ref{fig:large}, it can be easily observed that the proposed OMVFS methods achieves better performance than FSDS in most cases.
Especially when the number of selected feature is small (\textit{i.e.}, less than 300), 
OMVFS is much better than FSDS.
For example, on YouTube data, when we only select 100 features, 
the NMI for OMVFS is about 0.58, while the NMI for FSDS is only 0.28.
This observation suggests that unlike FSDS, which tends to include less discriminative features in the top selected feature set,
the proposed OMVFS selects more discriminative features even in a small set of selected features.

\subsection{Scalability Comparison} 
\begin{figure}[t]
	\centering
	\subfloat[Runtime on Reuters]{\includegraphics[width= .49\columnwidth]{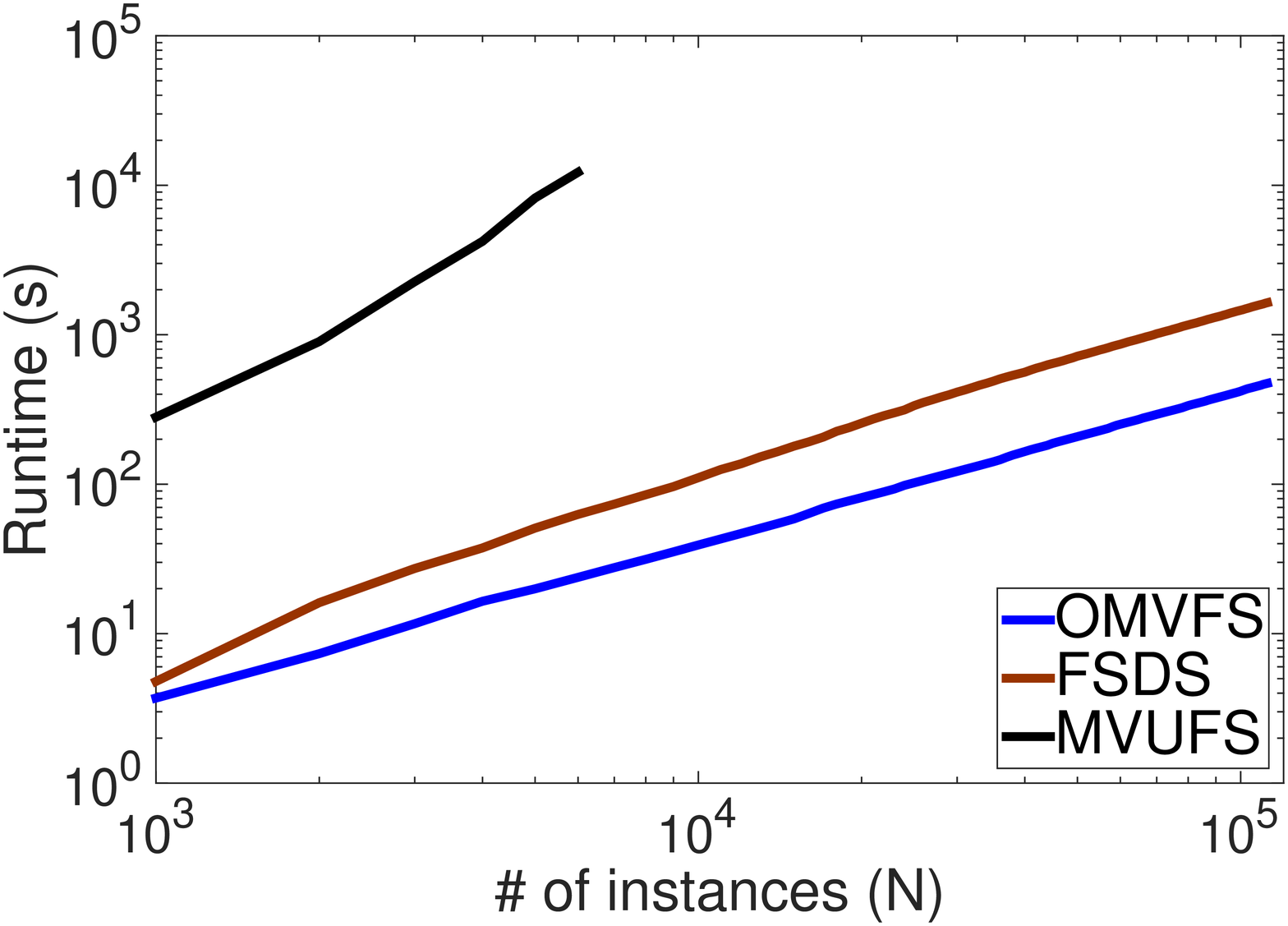}}
	\subfloat[Runtime on YouTube]{\includegraphics[width= .49\columnwidth]{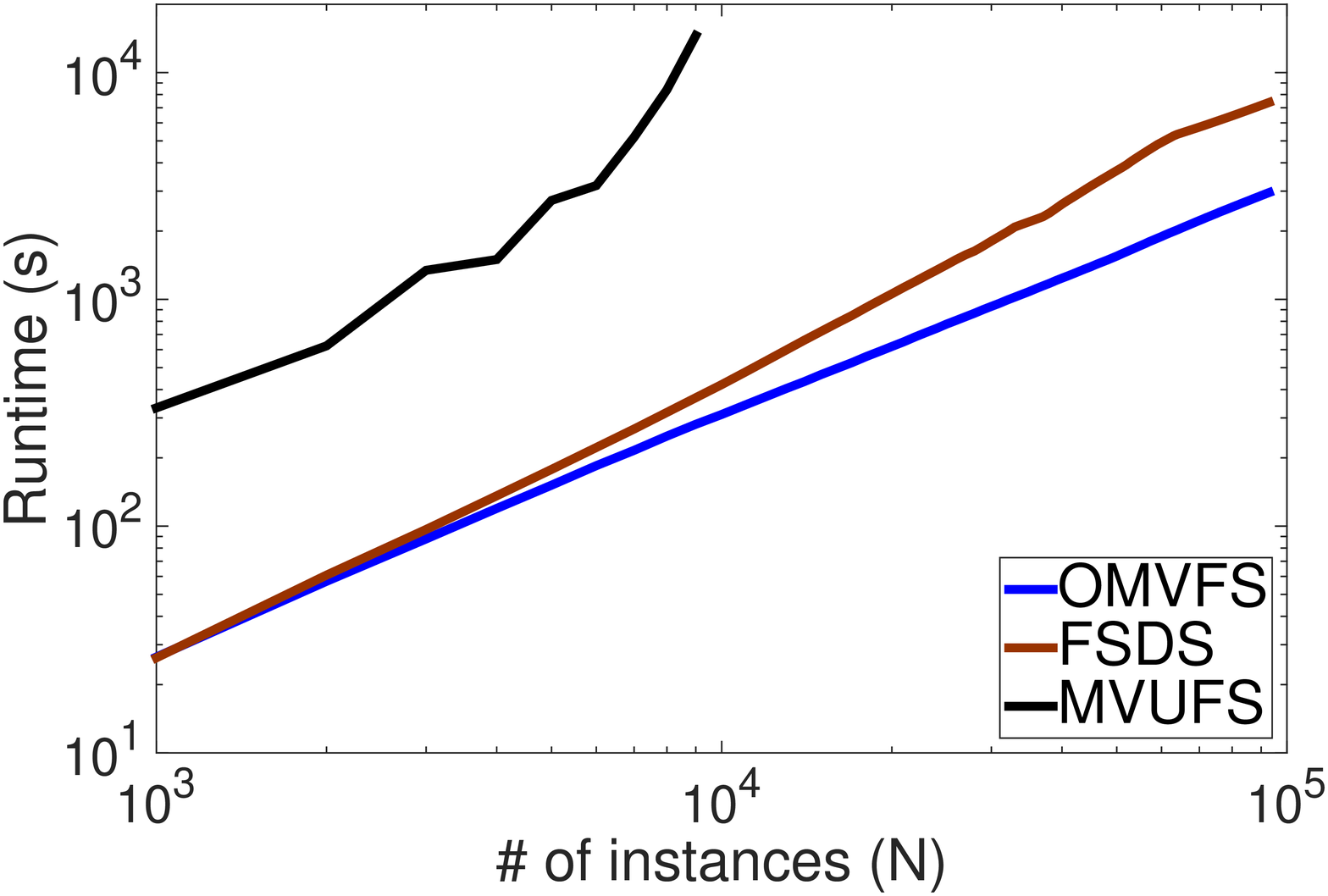}}
	\caption{Runtime v.s. the Data size on Reuters and YouTube Data. 
		We only report the runtime for MVUFS under size $ 10^4 $, because of the high runtime and memory consumption. }
	\label{fig:runtime}
\end{figure}

\begin{figure}[t]
	\centering
	\subfloat[Runtime on Reuters]{\includegraphics[width= .49\columnwidth]{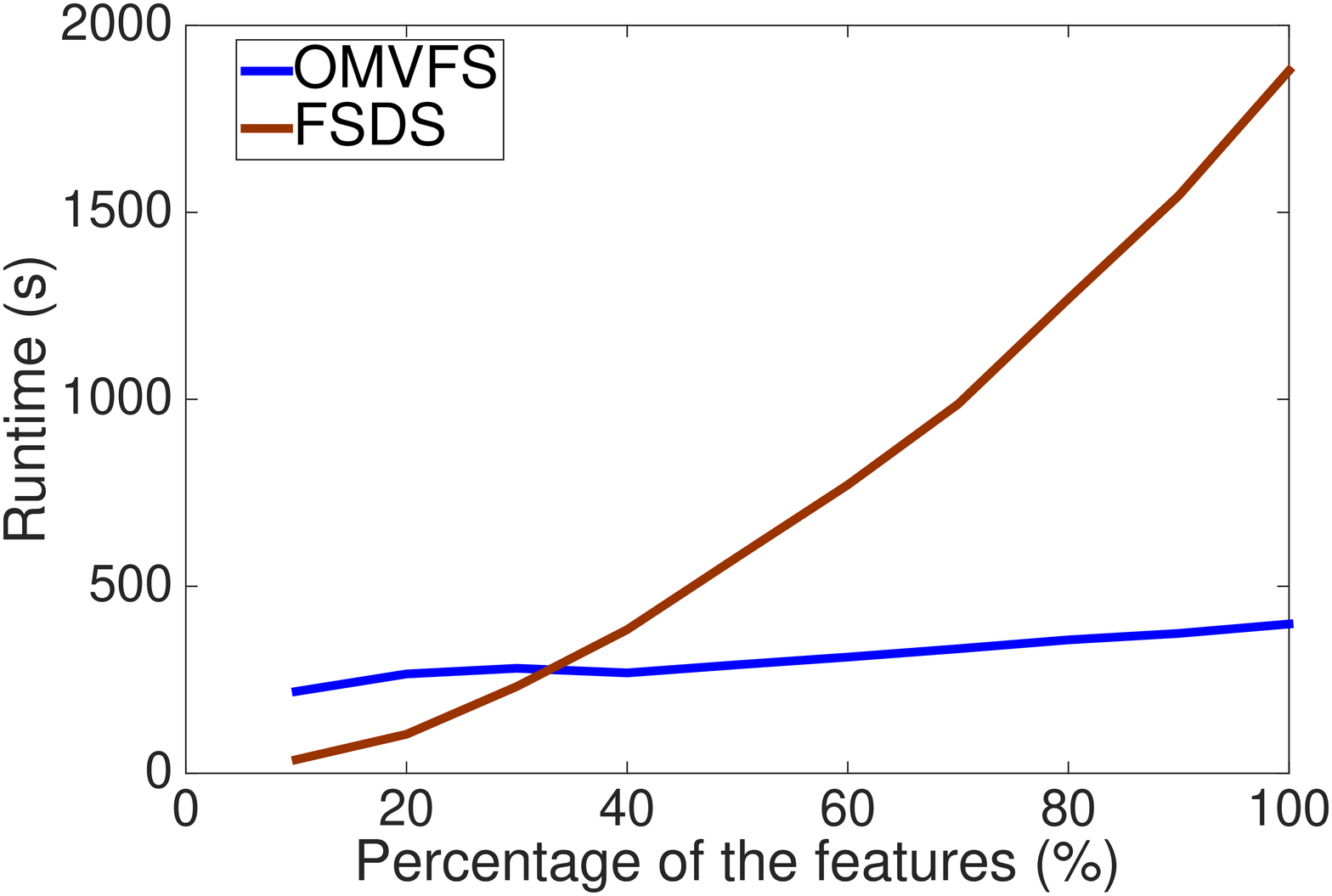}}
	\subfloat[Runtime on YouTube]{\includegraphics[width= .49\columnwidth]{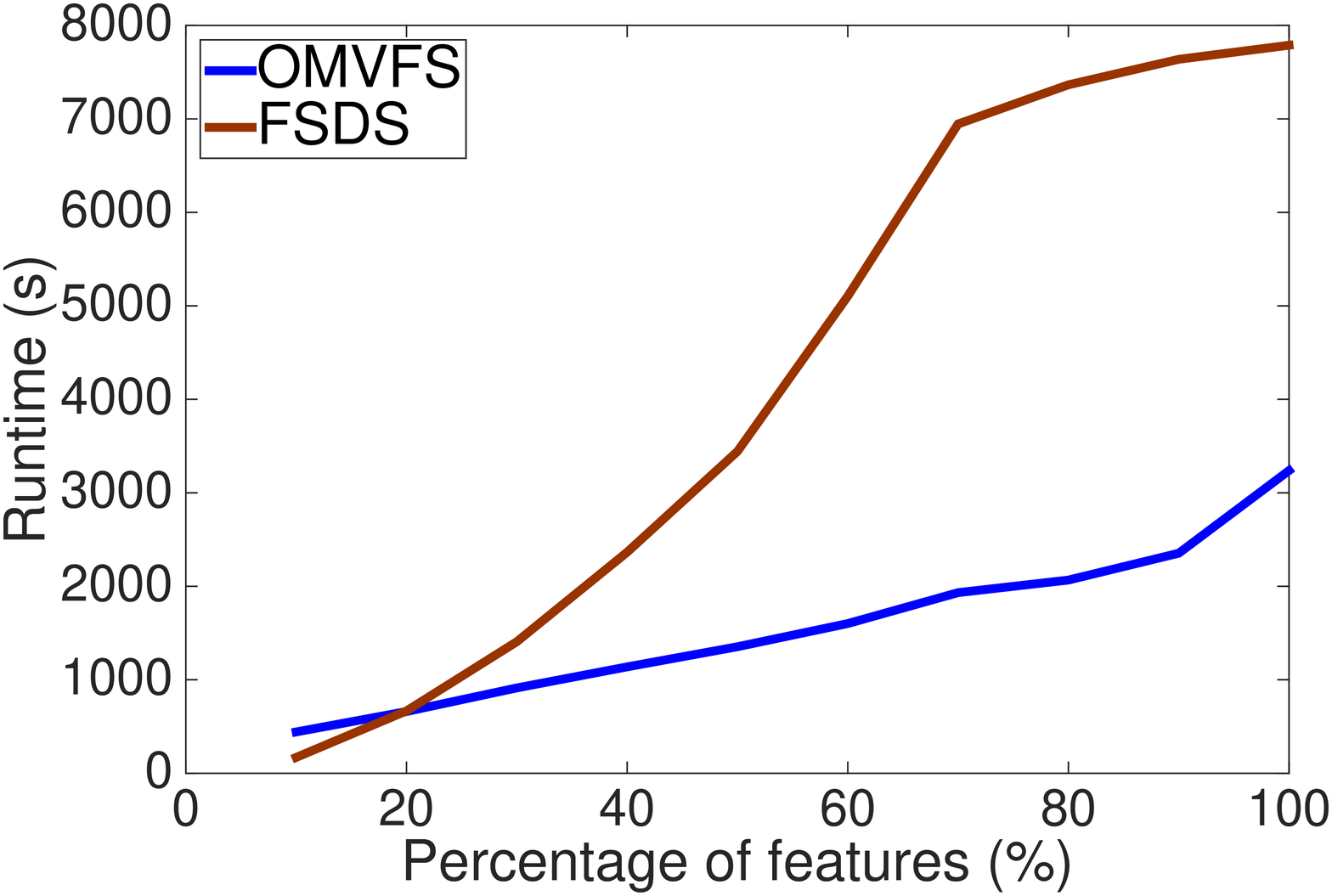}}
	\caption{Runtime v.s. the feature dimension on Reuters and YouTube Data. 
	Due to the high runtime and memory consumption of MVUFS, we only report the runtime for OMVFS and FSDS.
		}
	\label{fig:runtime_feature}
\end{figure}

In order to show the scalability of the proposed OMVFS method under different data sizes, 
we run OMVFS on the Reuters and YouTube data 
and report the runtime for different data sizes (number of instances) in Fig.~\ref{fig:runtime}.
Also, to show the scalability of OMVFS method under different feature dimensions, 
we randomly sample different numbers of features from the two large datasets and run OMVFS on the sampled features.
The runtime for different feature dimensions is reported in Fig.~\ref{fig:runtime_feature}.

We select MVUFS and FSDS as comparison methods, 
since MVUFS is the only one that deals with multi-view data 
and FSDS is the only one that handles large-scale/streaming data.
Because FSDS is a single view method,
we run FSDS on each view and report the total runtime for all the views.

From Fig.~\ref{fig:runtime}, we can observe that 
the two online methods, FSDS and OMVFS, are much faster than the off-line MVUFS and the runtime of the two online methods are linear with respect to the data size.
It only takes about 3 seconds for FSDS and OMVFS on data with size 1000 on Reuters data, while MVUFS takes about 300 seconds.
Like the FSDS method, the proposed OMVFS only uses the current data chunks 
and all the historical data is aggregated. 
OMVFS considers all the views simultaneously while FSDS can only work on single view.
Although the two online methods have similar runtime with small data, 
the proposed OMVFS uses less time as the data size increases. 
Further, it can be found that OMVFS is about 100 times faster than MVUFS on both datasets. 

Because of the high memory usage and slow runtime of MVUFS, we only report the runtime for OMVFS and FSDS in Fig.~\ref{fig:runtime_feature}. 
From Fig.~\ref{fig:runtime_feature}, we can observe that although FSDS is faster when we only sample small number of features (10 \% or 20\%),
the runtime of OMVFS grows much slower than FSDS as the dimension of features increases. 
For Reuters data, the runtime of OMVFS only increases from 240 seconds to 480 seconds when the number of features increases from 10\% to 100\%.
Thus, the proposed OMVFS is faster when the feature dimension becomes really large.

\subsection{Stability under Concept Drift}
\begin{figure}[t]
	\centering
	\includegraphics[width=0.7\columnwidth]{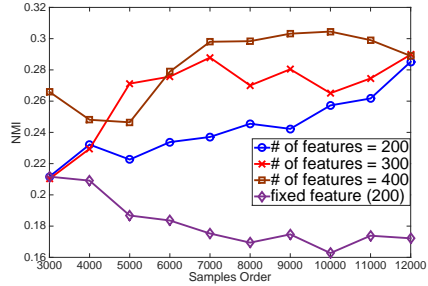}
	\caption{Concept drift test across the data stream for OMVFS}
	\label{fig:concept_drift}
\end{figure}

It is well-known that online/streaming algorithms are generally sensitive to the order of data, or concept drift \cite{wang2003mining}. 
To test the performance of OMVFS in such scenarios, we use 12,000 instances from Reuters data.
We randomly create unbalanced data with two dominant classes and randomly change the dominant classes for every 3,000 instances in the data stream.
OMVFS is then applied to the data stream with concept drifts. 
As the data came in, we report the performance of OMVFS with different sizes of feature set in Fig.~\ref{fig:concept_drift}. 
We also compare against a scheme where  a static feature subset (with 200 features) is used without adapting to concept drift.
This static feature subset is determined by OMVFS using only the first 3,000 instances in the data stream.

From Fig.~\ref{fig:concept_drift}, we can observe that the approach based on static feature subset performs quite close to OMVFS in the beginging.
However, as new data come in and concept drift becomes more prominent,
the performance of static feature subset decreases.
On the other side, the performance of the proposed OMVFS will not decrease as more data come in. 
Instead, the performance increases as OMVFS combines the information in the new data with the aggregated information from previous data.
Also, from Fig.~\ref{fig:concept_drift}, we can see that in general, 
the more features we select, the better performance OMVFS would achieve.

\subsection{Parameter Study}
\begin{figure}[t]
	\centering
	\subfloat[ACC when $ \beta_v = 1$, with different $ \alpha_v $]{\includegraphics[width= .455\columnwidth]{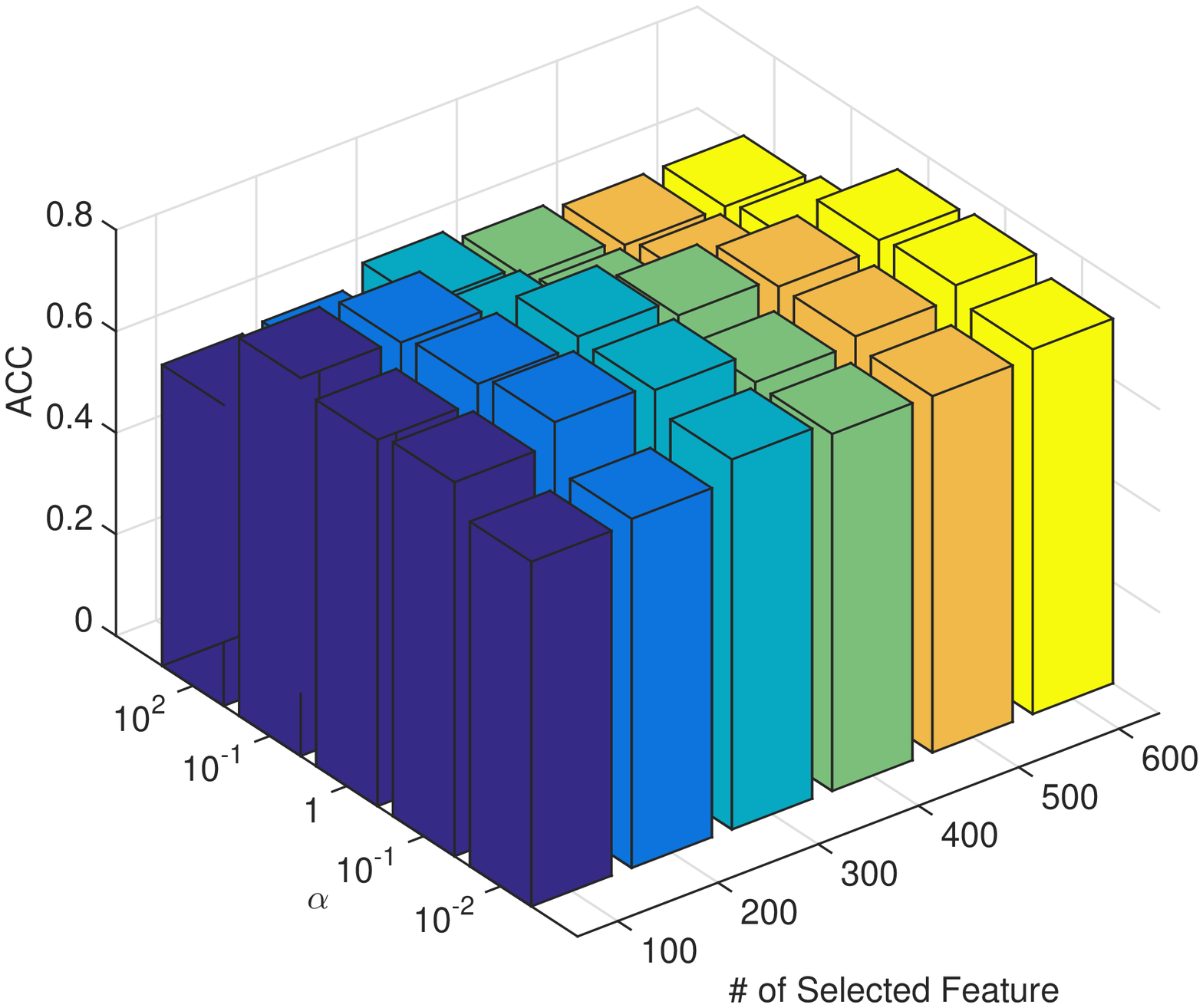}}
	~
	\subfloat[ACC when $ \alpha_v = 10$, with different $ \beta_v $]{\includegraphics[width= .535\columnwidth]{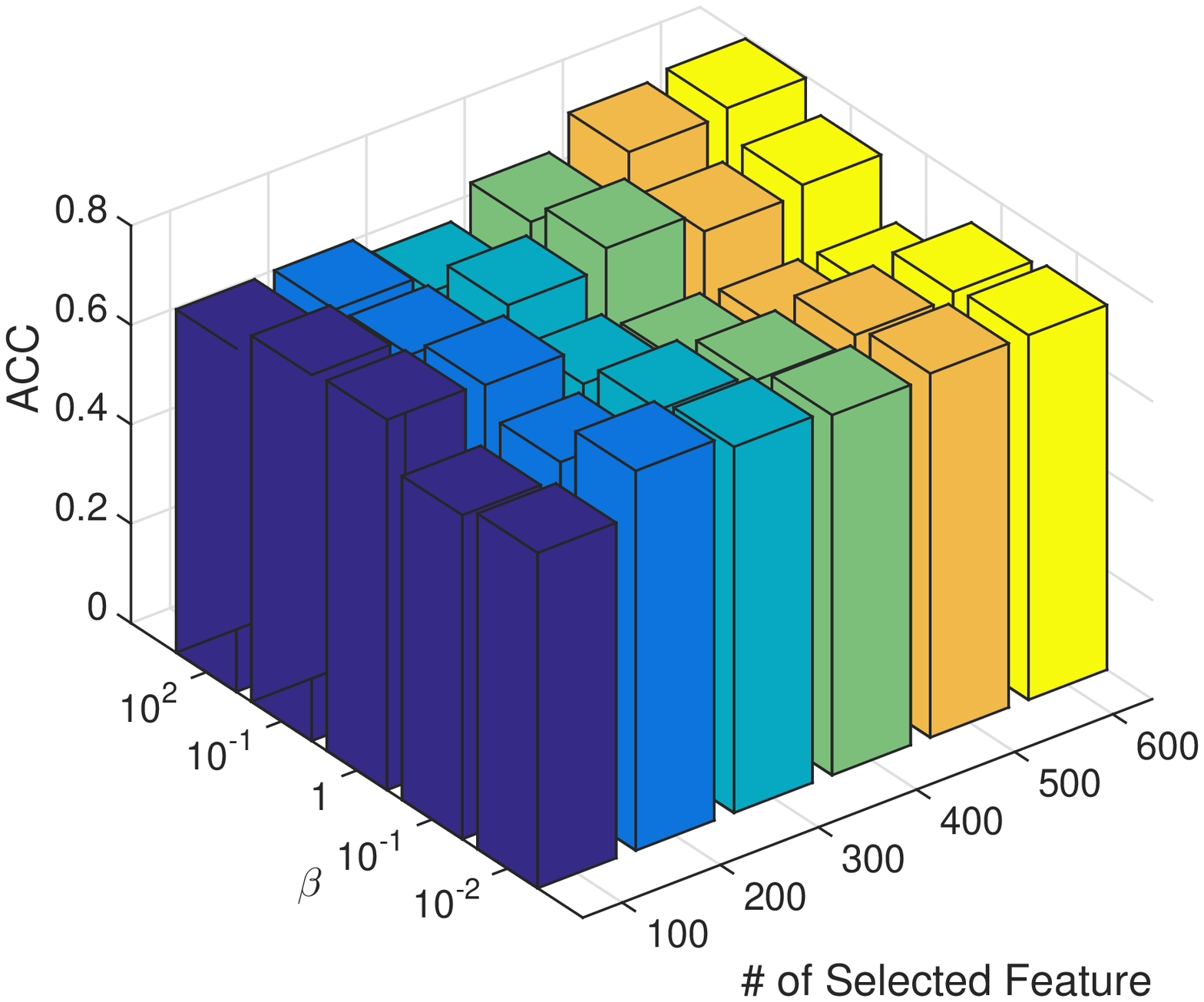}}
	\\
	\subfloat[NMI when $ \beta_v = 1$, with different $ \alpha_v $]{\includegraphics[width= .455\columnwidth]{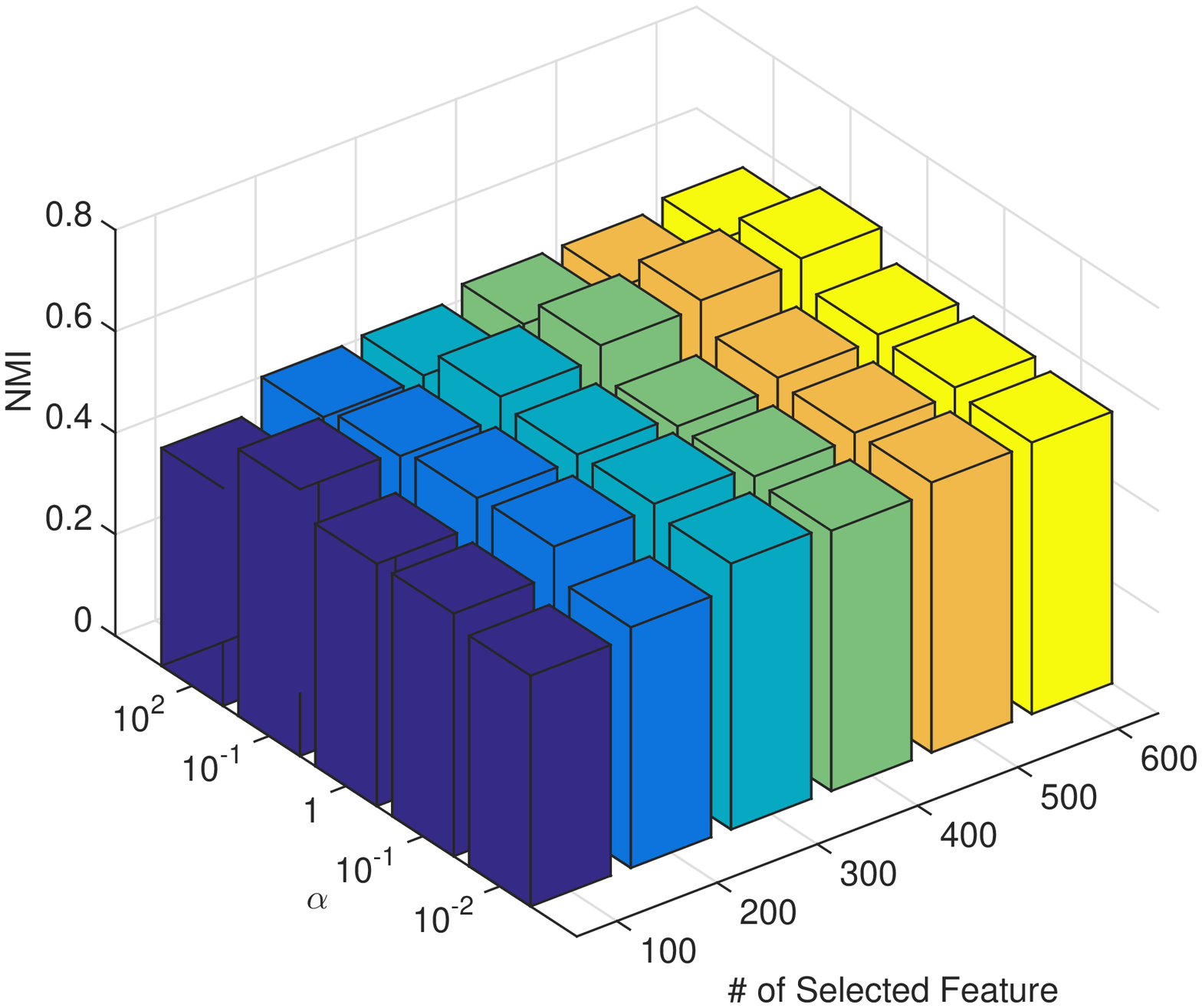}}
	~
	\subfloat[NMI when $ \alpha_v = 10$, with different $ \beta_v $]{\includegraphics[width= .535\columnwidth]{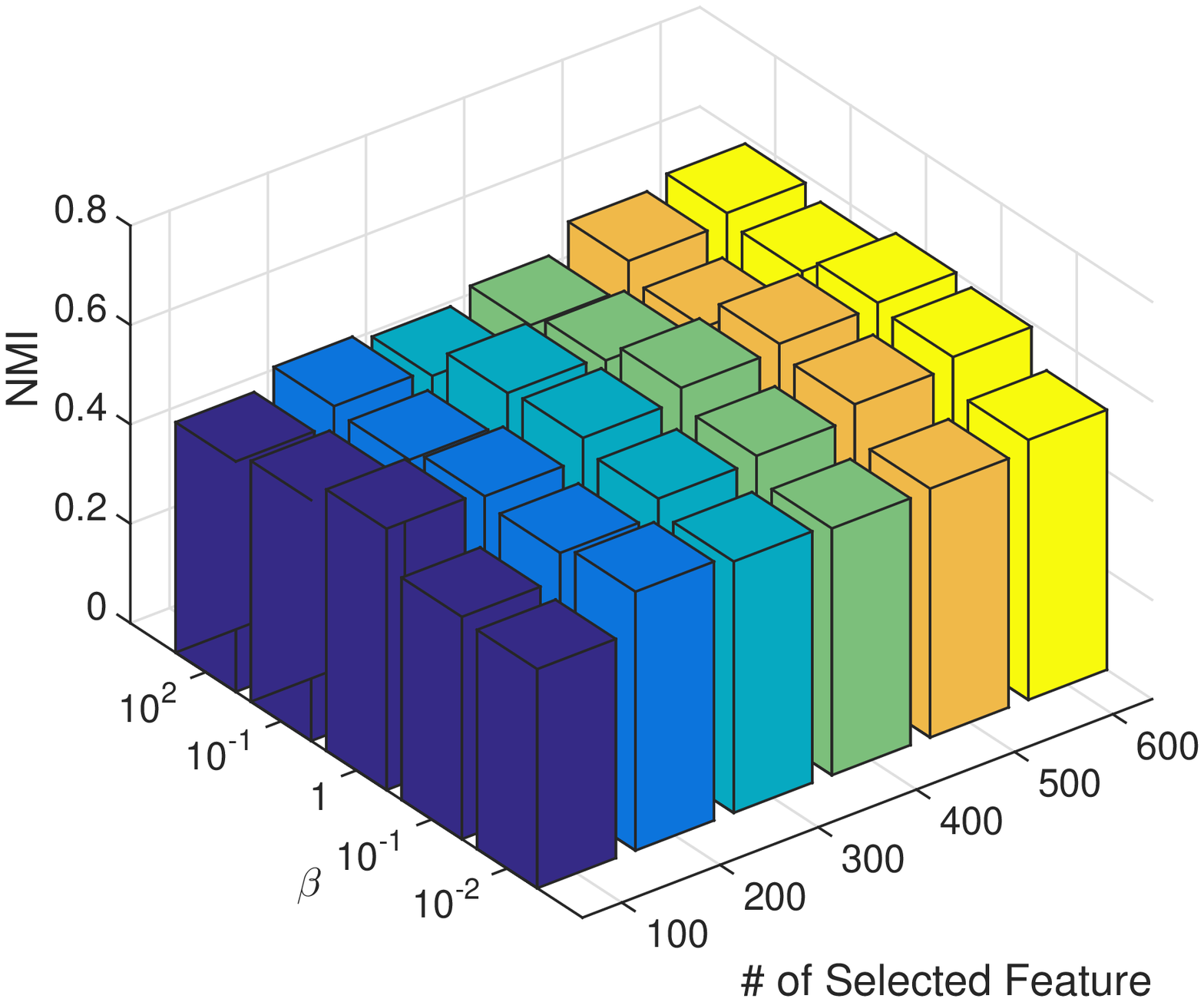}}
	\caption{ACC and NMI of OMVFS with different $ \alpha_v $,$ \beta_v $ and feature numbers on FOX}
	\label{fig:parameter}
\end{figure}

There are two sets of parameters in the proposed methods: $ \{\alpha_v\} $ and $ \{\beta_v\} $.
Here, we explore the effects of the two parameter sets.
For the sake of convenience, we set $ \alpha_v $ to be equal for different views and also set $ \beta_v $ equally.
We ran OMVFS with different values for $ \{\alpha_v\} $ and $ \{\beta_v\} $ on FOX data. 
Basically, we fix one of the parameters and ran OMVFS with different values for the other.
We report the performance with different sizes of the selected features.
The results in ACC and NMI are shown in Fig.~\ref{fig:parameter}.

From Fig.~\ref{fig:parameter}, we can see that for most of the cases,
the proposed OMVFS method is not very sensitive to the parameters $ \alpha_v $ and $ \beta_v $.
However, the performance does increase as the number of selected features increases.

\begin{figure}
	\centering
	\subfloat[]{\includegraphics[width= .48\columnwidth]{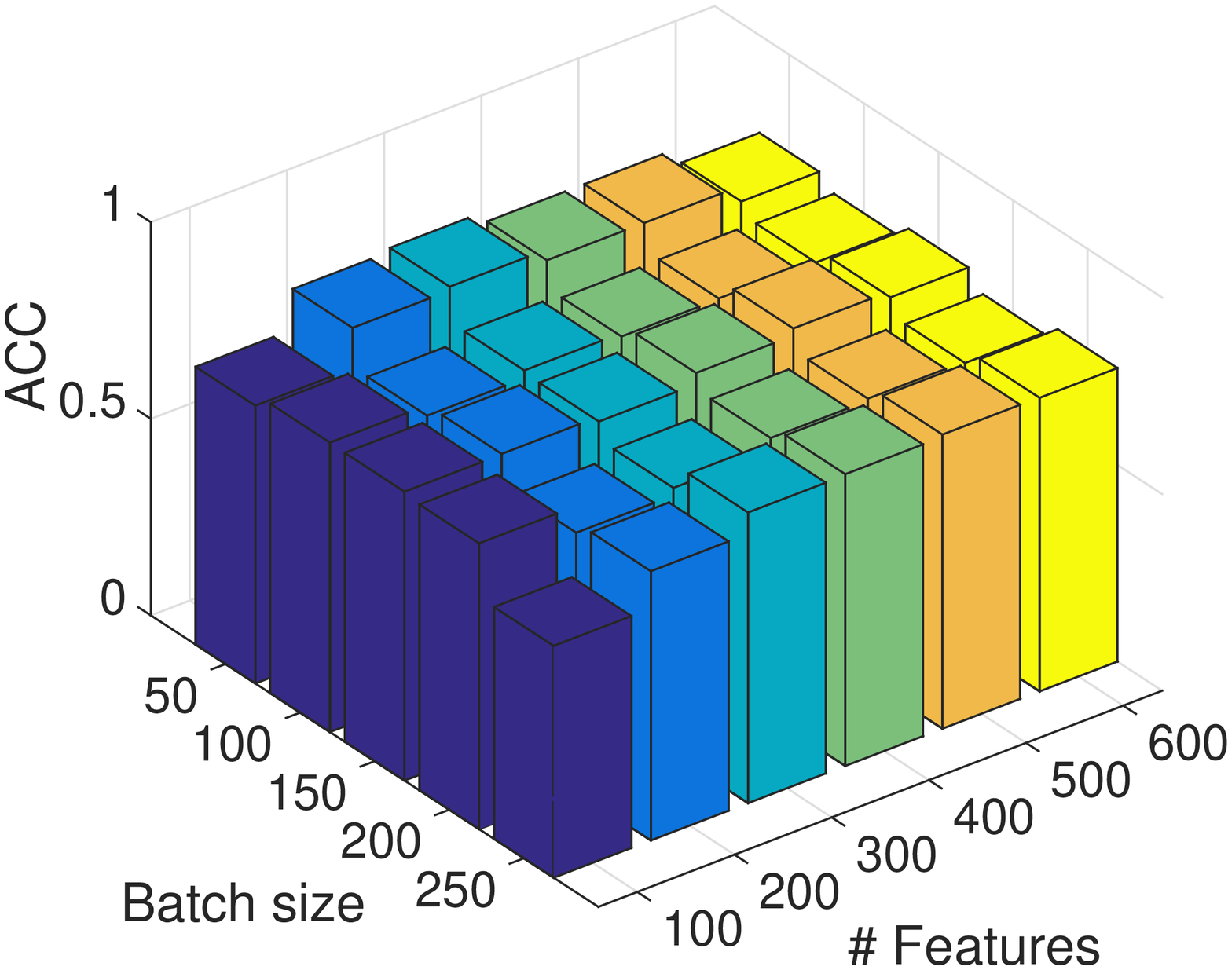}}
	~
	\subfloat[]{\includegraphics[width= .48\columnwidth]{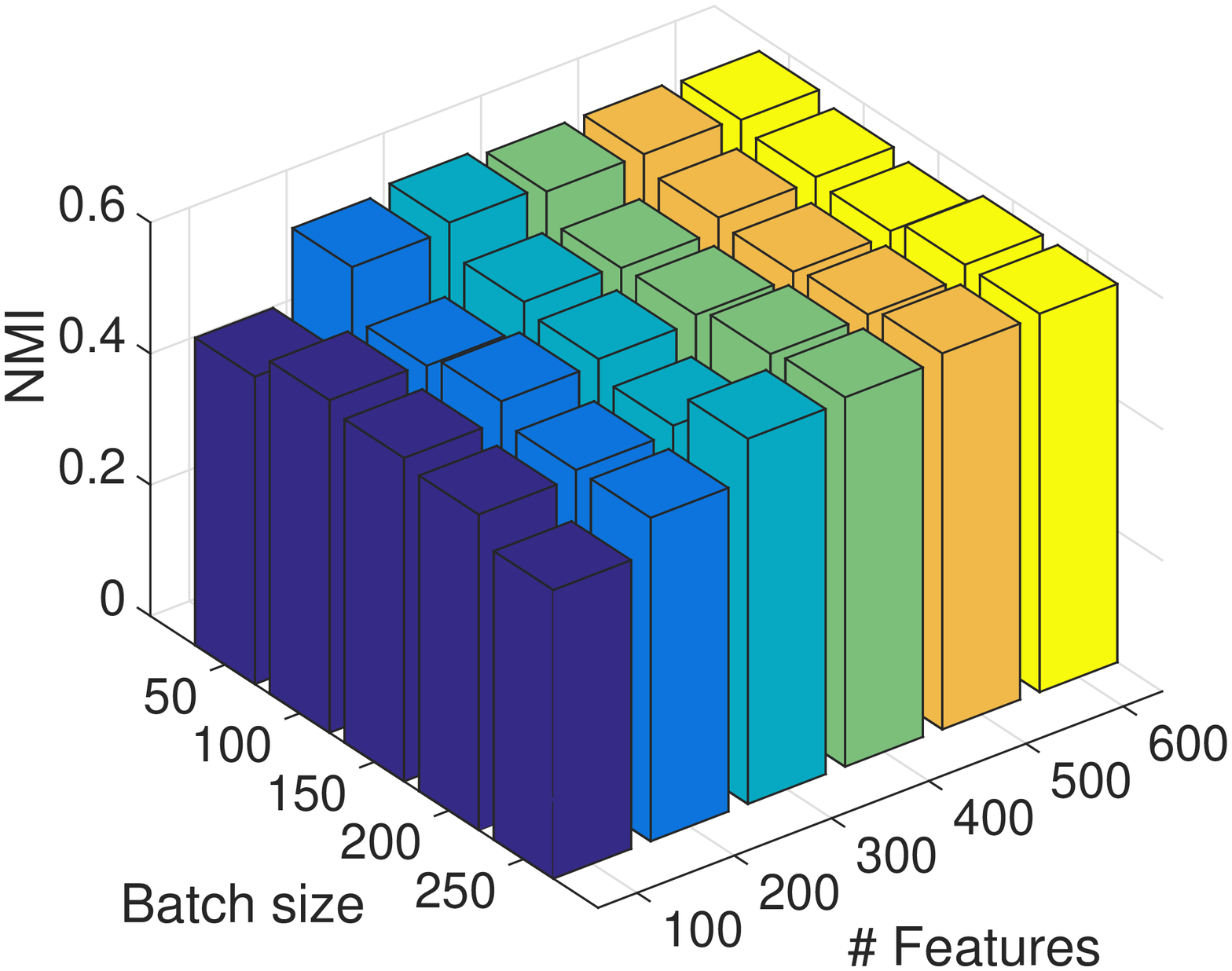}}
	\caption{ACC and NMI of OMVFS with different batch sizes $ m $ and feature numbers on FOX}
	\label{fig:batch_size}
\end{figure}

Another parameter in the proposed OMVFS method is the batch size $ m $, which is a common parameter for streaming algorithms.
To test the performance of OMVFS under different batch sizes,
we run OMVFS on FOX data with different batch sizes and numbers of selected features.
The results in ACC and NMI are shown in Fig.~\ref{fig:batch_size}.
From Fig.~\ref{fig:batch_size}, we can clearly observe that the performance of OMVFS is very stable under different batch sizes.

\section{Related Work}
\label{sec:related}
There are three areas of related works upon which the proposed model is built.
Feature selection, especially unsupervised feature selection \cite{cai2010unsupervised,wang2015embedded}, is the first area that is related to this work.
Most of the unsupervised feature selection methods combine generated pseudo labels with sparse learning \cite{li2012unsupervised,qian2013robust}.
In terms of the number of views available, unsupervised feature selection can be categorized into single view feature selection and multi-view feature selection.
Most of the conventional unsupervised feature selection methods are designed for single view. 
As more and more multi-view data are generated, the limitation of conventional feature selection methods becomes more obvious.
Multi-view feature selection has drawn more attention in recent years.
Several effective methods have been proposed to solve unsupervised feature selection problem for various multi-view scenarios \cite{wang2013multi,tang2013unsupervised,qian2014unsupervised}.
However, all the previous multi-view feature selection methods suffer from the scalability issues. 
They all require that the whole data fit into the memory. 
The proposed OMVFS method, however, only takes a small amount of memory space and uses less computational time without sacrificing the performance, which makes it more suitable for large-scale/streaming data.

Multi-view unsupervised learning is the second area that is related to our work. 
A few directions were explored to solve multi-view unsupervised learning in recent years.
For example, \cite{mvc_cca09_cp, mvc_cokl2013} propose to use canonical correlation analysis to combine different views.
\cite{sdm2013_liu, mic_ecml} explore the option to model the multi-view learning as a joint NMF problem.
\cite{liu2013multiview, shao2015clustering} use tensor to model the multi-view data. 

Nonnegative matrix factorization \cite{lee2001algorithms}, 
especially online NMF, is the third area that is related to our work.
NMF has been successfully used in unsupervised learning \cite{cai2011graph,ding2006orthogonal}.
However, traditional NMF cannot deal with large-scale data. 
Different variations were proposed in the last few years.
For example, \cite{wang2011efficient} proposed an online NMF algorithm for document clustering. 
\cite{guan2012online} proposed an efficient online NMF algorithm (OR-NMF) 
that takes one sample or a chunk of samples per step and updates the bases via robust hastic approximation.
All the online NMF methods either focus on clustering or dimension reduction. 
None of them are designed for feature selection.
Furthermore, none of them can handle multi-view data.
However, our proposed OMVFS directly embeds feature selection into the online joint NMF framework with graph regularization,
which handles large/streaming multi-view data.

\section{Conclusions}
\label{sec:conclusion}
In this paper, we present possibly the first attempt to solve the online unsupervised multi-view feature selection problem.
Based on NMF, the proposed method OMVFS directly embeds the feature selection into the graph regularized clustering algorithm.
A joint NMF is used to learn a consensus clustering indicator matrix, which makes OMVFS integrate information from different views.
OMVFS also adopts the graph regularization to preserve the local structure information and help select more discriminative features.
Solving the optimization problem in an incremental way, OMVFS can process the data chunks one by one without storing all the historical data, which greately reduces the memory requirement. 
Also, by using the buffering technique, OMVFS can reduce the computational and storage cost while taking advantage of the structure information.
The buffering will also help OMVFS capture the concept drift in the data streams.
Extensive experiments conducted on two small datasets and two large datasets demonstrate the effectiveness and efficiency of the proposed OMVFS algorithm.
Without sacrificing performance, the proposed OMVFS is about 100 times faster than the best off-line multi-view feature selection method.




\bibliographystyle{IEEEtran}
\bibliography{mybib}
%



\end{document}